\documentclass[10pt,twocolumn]{article}
\usepackage[letterpaper,margin=0.70in,columnsep=0.25in]{geometry}
\usepackage[T1]{fontenc}
\usepackage{amsmath}
\usepackage{newtxtext,newtxmath,microtype}
\usepackage{graphicx,booktabs,caption,xcolor,enumitem,titlesec,xurl}
\usepackage{flushend,balance}
\usepackage[numbers,sort&compress]{natbib}
\usepackage[ruled,vlined]{algorithm2e}
\usepackage[colorlinks=true,allcolors=linkblue]{hyperref}
\definecolor{linkblue}{HTML}{2E5B83}
\definecolor{cardinal}{HTML}{8C1515}
\hypersetup{pdftitle={Can LLMs Debug Robots by Editing Their Weights?},pdfauthor={Jusheng Zhang; Keze Wang; Liang Lin}}
\titleformat{\section}{\large\bfseries}{\thesection}{.55em}{}
\titleformat{\subsection}{\normalsize\bfseries}{\thesubsection}{.55em}{}
\titleformat{\paragraph}[runin]{\bfseries}{}{0pt}{}
\titlespacing*{\section}{0pt}{8pt plus 1pt minus 1pt}{4pt}
\titlespacing*{\subsection}{0pt}{6pt plus 1pt minus 1pt}{3pt}
\titlespacing*{\paragraph}{0pt}{5pt}{.5em}
\setlist[itemize]{leftmargin=*,topsep=3pt,itemsep=2pt,parsep=0pt}
\SetAlFnt{\small}
\SetAlCapFnt{\small\bfseries}
\SetAlCapNameFnt{\small\bfseries}
\newcommand{\pp}{\,\mathrm{pp}}
\newcommand{\code}[1]{\texttt{\small #1}}


\begin{document}
\makeatletter\if@filesw\immediate\write\@auxout{\string\citation{bstcontrol}}\fi\makeatother
\twocolumn[{
\begin{minipage}{\textwidth}
\includegraphics[width=.95in]{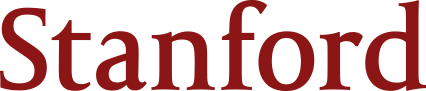}\hfill{\footnotesize November 6, 2026}
\vspace{8pt}
\begin{center}
{\LARGE\bfseries Can LLMs Debug Robots by Editing Their Weights?\par}
\vspace{6pt}
{\large \href{https://jushengzhang.github.io/}{Jusheng Zhang}$^{1}$\qquad \href{https://kezewang.com/}{Keze Wang}$^{2}$\qquad \href{https://scholar.google.com/citations?user=Nav8m8gAAAAJ&hl=zh-CN}{Liang Lin}$^{2}$\par}
\vspace{3pt}
{$^{1}$Stanford University\qquad $^{2}$X-Era AI\par}
\end{center}
\vspace{4pt}
\noindent\makebox[\textwidth][c]{\includegraphics[width=\textwidth]{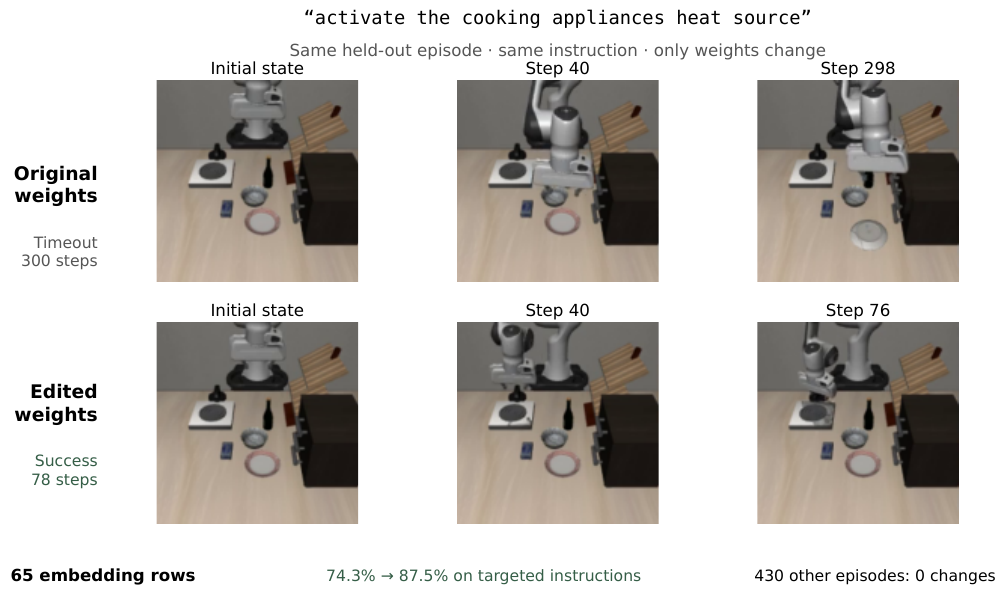}}
\captionof{figure}{\textbf{A 65-row weight edit changes which task a robot performs.} On held-out LIBERO-Goal variant 1421 (seed 7, initial state 0), original $\pi_{0.5}$ picks up the plate and times out at step 300; run C turns the stove knob and reaches the goal at step 78. The instruction is unchanged. Columns show initial, step-40, and each trajectory's last recorded camera frame (298/76); recording precedes the final success/timeout check. The metric strip summarizes all held-out instructions containing edited tokens, not this selected episode alone.}
\label{fig:teaser}
\vspace{9pt}
\end{minipage}
}]
\begin{abstract}
A language model can repair a robot policy's lexical instruction grounding by editing just 65 token-embedding rows. We bring weight-space programming to robot foundation policies: GPT-6 Astra reads weights and uses a CPU workspace and budgeted simulator feedback, without gradient updates or new demonstrations. Three independent sessions on $\pi_{0.5}$ all improve success on 765 held-out paraphrases; the best raises success from 86.3\% to 92.0\% (48 repairs, four regressions; exact paired $p=1.31\times10^{-10}$). Among the 335 instructions containing edited tokens, success rises from 74.3\% to 87.5\%; the other 430 episodes have zero outcome changes. Matched random-direction edits do not help. Full replacement using the editor's own lexicon explains much of the gain but produces more regressions; its direct difference from the editor is not significant at 0.05. Across 13 further robot sessions, the editor declines 12 candidates; the sole retained edit does not show a significant held-out gain. Separately, benchmark metadata lowers FastWAM success from 70.0\% to 48.6\%, while the evaluated VLA's aggregate rate is unchanged. The evidence supports precise lexical repair, not general acquisition of missing robot capabilities.
\end{abstract}
\section{Introduction}
A language model can repair some robot-policy failures by changing a few rows of the policy's weights. Our strongest result changes 65 token-embedding rows: success rises from 74.3\% to 87.5\% on 335 held-out instructions containing an edited token, with no outcome changes on the other 430 instructions. The manipulation skill is already available; the edit changes which language reaches it.

Consider a policy that turns on the stove under its canonical instruction, but picks up a plate when asked to ``activate the cooking appliances heat source'' (Figure~\ref{fig:teaser}). Both actions are familiar tasks in the same scene. The failure is a mismatch between wording and the task represented by the policy, rather than the absence of a stove-turning behavior. A camera perturbation or a dropped object presents a different debugging problem: changing the interpretation of a word may not supply the required visual or recovery behavior.

Takehi et al.~\cite{wsp} introduced \emph{weight-space programming}: a general-purpose LLM inspects and changes another network's parameters directly, without gradient optimization. Their demonstrations span games and language models, and motivate a distinction between removing an obstacle to an existing ability and creating a new ability. We bring this experimental setup to robot foundation policies. We test both whether an editor can find a useful intervention and whether it declines interventions that fail its own checks.

GPT-6 Astra receives policy weights, a persistent CPU Python workspace, and a budgeted forward tool. It can inspect tensors and write replacements, but cannot train the policy or change the deployment program. Practice rollouts and probes guide editing; held-out variants are unavailable to the editor. Across three independent sessions on $\pi_{0.5}$~\cite{pi05}, Astra converges on sparse lexical remapping. All three submitted edits improve held-out paraphrase success. The best changes less than 0.004\% of exported parameters and raises overall success from 86.3\% to 92.0\% on all 765 held-out language variants.

The boundary matters as much as the gain. In 13 additional sessions over FastWAM~\cite{fastwam}, LIBERO-Plus~\cite{liberoplus}, LIBERO-PRO~\cite{liberopro}, and LIBERO-RECOVER~\cite{liberorecover}, the editor declines 12 final candidates. The sole retained candidate has no significant held-out gain. Two archived, declined candidates do have positive aggregate estimates; we report them rather than treating final-directory contents as accepted repairs. These observations support a limited, practical form of debugging, not a general impossibility theorem about acquiring capabilities.

Our contributions are:
\begin{itemize}
\item A reproducible lexical repair of a VLA: three independent editing sessions, one fixed held-out population, and outcome changes confined to the edited-token instructions.
\item Controls that separate direction from edit size and full replacement from the editor's interpolation, together with an explicit account of what a dictionary control already explains.
\item A 16-session record of retained and declined edits, plus a separate finding that benchmark instruction metadata strongly affects FastWAM but not the evaluated $\pi_{0.5}$ checkpoint in aggregate.
\end{itemize}

\section{A Fixed Interface for Weight-Space Debugging}
\label{sec:protocol}
\begin{figure*}[t]
\centering
\includegraphics[width=\textwidth]{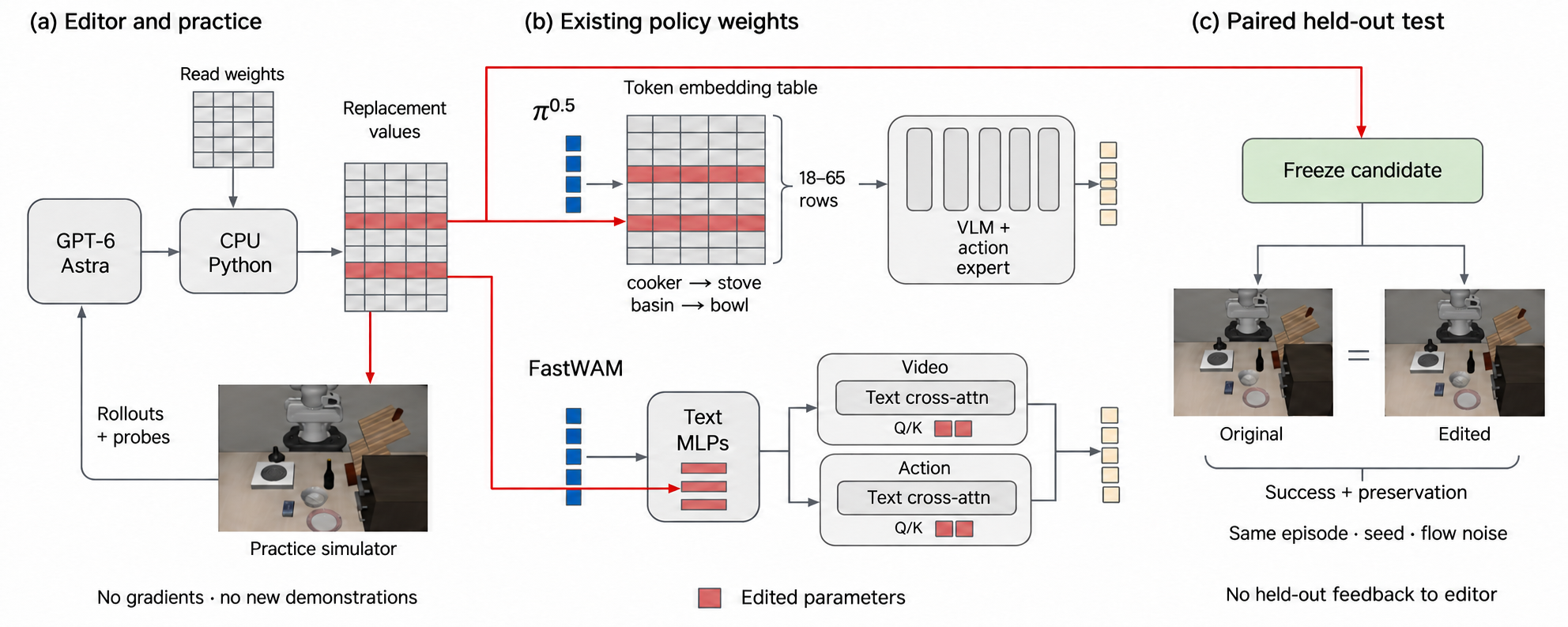}
\caption{\textbf{One editing interface, two policy architectures.} (a) Astra reads weights and writes replacement tensors in a CPU kernel; a separate simulator provides practice rollouts and replay probes. (b) The successful $\pi_{0.5}$ edits change token-embedding rows. The FastWAM investigation changes text-conditioning MLP and Q/K parameters. These are replacements within existing modules, not newly inserted runtime components. (c) Candidate values are frozen before paired held-out evaluation, whose outcomes are not returned to the editor. Red denotes parameter locations. Layout follows the supplied GR00T N1 visual reference~\cite{grootn1}; camera insets are illustrative.}
\label{fig:workflow}
\end{figure*}
\paragraph{Weights, tools, and constraints.}
Figure~\ref{fig:workflow} shows the common interface. The editor receives tensor names and shapes, read access to original weights, a tokenizer where applicable, and a CPU Python kernel whose state persists between calls. It can perform static linear algebra and save or remove replacement tensors. The protocol forbids gradient updates, training loops, iterative loss fitting, and execution of the target network inside the kernel. The model is evaluated with gradients disabled. Only weight values may change: tensor names, shapes, input instructions, action scaling, and inference code remain fixed.

A separate forward tool runs simulator episodes or replays saved model inputs for action and activation probes. The kernel can inspect returned observations and outcomes. Thus \emph{no additional demonstrations} does not mean no observations or feedback; the policies were already pretrained and fine-tuned on demonstrations. A forward request is a batch, not one neural-network evaluation. Robot sessions receive 40--50 requests; all three language-repair sessions receive 50. A Python execution may use up to 900 seconds and 96 GB of CPU memory. No extra policy training is performed for this report.

\paragraph{Splits and candidate selection.}
Practice and held-out variants are separated by the parity of a SHA-256 hash of suite and variant ID. The language task exposes 4,986 practice variants across perturbation categories, but no held-out list. Each session reserves its own practice validation before making a final retain/revert decision. The three $\pi_{0.5}$ sessions start separately with the same checkpoint and task specification; they do not inherit each other's notes or edits. They are independent searches, not independent test datasets. In contrast, the three FastWAM suffix follow-ups inherit the first session's notes and archived candidate.

After editing, each submitted language candidate is evaluated on all 765 held-out Language Instructions variants and a 360-episode sample of the other six categories. The latter sample is shared across sessions. A smaller, earlier sample of 200 held-out language variants is disclosed as well (Appendix~\ref{app:language}); it did not support a significant gain. The all-variant population is fixed by the original split, not selected from successful examples, and the candidates are not tuned on either held-out evaluation.

\paragraph{Pairing and statistics.}
Original and edited weights are compared on the same variant, seed, and initial state. Held-out language evaluation uses seed 7. Weight overrides are applied in place; restoration is verified afterward. Let $g$ count original failures that become successes and $l$ original successes that become failures. We report $(g-l)/n$, paired normal-approximation intervals, and exact two-sided McNemar tests~\cite{mcnemar}. Tests refer to the evaluated episode pairs, not uncertainty over editor sessions. Category and follow-up comparisons are exploratory. The three primary language effects also remain significant under Holm correction within that three-test family. Runtime-error handling and complete counts are in Appendix~\ref{app:language}.

\paragraph{What is and is not replicated.}
Pairing measures changed outcomes, not just two independent success rates: C has 52 discordances, of which 48 favor the edit. Its exact test conditions on those discordances. The three language effects remain significant after Holm correction (adjusted $p=3.92\times10^{-10}$, $3.64\times10^{-4}$, and .0106 for C, A, and B). This correction covers the three primary candidates, not every exploratory probe in the study. Independent sessions test whether the editor can find an edit again; shared variants and one held-out seed limit claims about environmental generalization.

The all-variant evaluation follows the weaker 200-episode check. Its population is the entire pre-existing held-out language split, and the candidates remain unchanged. This avoids selecting variants that respond well to an edit, but does not turn the larger test into a preregistered independent replication. We disclose both evaluations and their overlap. The complete split, rather than only successful episodes or hard examples, defines the reported 765-pair result.

\subsection{Checking the motivating language-model edit}
Before robot editing, we reconstruct the released Qwen3.5-0.8B \code{p50c} edit from Takehi et al.~\cite{wsp}, verifying the original and edited tensor hashes. With one sample per MATH-500 problem, it scores 49.4\%, compared with the report's 48.1\% using four samples. The local unedited checkpoint scores 9.0\% on a fixed 100-problem subset, versus the reported 8.8\% on that subset. This is a reproduction of a released edit, not a new discovery by the robot editor. The subset baseline is not a matched full-500 comparison. Appendix~\ref{app:qwen} gives the protocol differences and truncation rates.

\section{Repairing a VLA's Instruction Grounding}
\label{sec:pi05}
\subsection{The skill exists; the wording fails}
We use openpi's released \code{pi05\_libero} checkpoint~\cite{pi05}. Its export contains 3.62 billion parameters: SigLIP vision encoding~\cite{siglip}, a PaliGemma-style VLM prefix~\cite{paligemma}, and a smaller flow-matching action expert. The evaluated LIBERO configuration generates ten future actions with ten denoising steps and executes five before replanning. The configuration disables discrete-state language input; this detail is checkpoint-specific. All settings are fixed during editing.

On a 700-episode practice scan, the policy succeeds on 86\% overall. A prominent weakness is paraphrasing in LIBERO-Goal, where ten tasks share a scene: success is 14/25 in the scanned cell. The policy often selects another familiar task or object when canonical names are replaced by synonyms or descriptions. All three sessions diagnose a content problem rather than a simple absence of attention to language. An action expert can attend to an instruction and still act on the wrong task representation.

The editor identifies task words such as ``cooker'', ``basin'', and ``fermented'' whose representations lead to the wrong behavior. These words need not be unknown to the pretrained language model; the operational contrast is with the checkpoint's canonical LIBERO task instructions. Each session protects the token IDs appearing in canonical instructions and remaps selected source-token embeddings toward task anchors such as stove, bowl, and wine. No vision, attention, or action-expert tensor changes in the submitted language edits.

\subsection{Three searches, three held-out improvements}
\begin{table}[t]
\centering\small
\setlength{\tabcolsep}{3.6pt}
\begin{tabular}{lrrrrr}
\toprule
Run & Rows & Original & Edited & $g/l$ & Exact $p$\\
\midrule
A & 18 & 86.3 & 89.4 & 32/8 & $1.82\!\times\!10^{-4}$\\
B & 20 & 86.3 & 88.0 & 18/5 & 0.0106\\
C & 65 & 86.3 & \textbf{92.0} & 48/4 & $1.31\!\times\!10^{-10}$\\
\bottomrule
\end{tabular}
\caption{\textbf{All three independent language-editing sessions improve held-out success.} Percentages on the same 765 held-out paraphrase variants; $g/l$ denotes repairs/regressions. No run is omitted. The common original count is 660; edited counts are 684, 673, and 704.}
\label{tab:pi05-main}
\end{table}
Table~\ref{tab:pi05-main} reports all three sessions. The strongest candidate, C, repairs 48 failures and breaks four successes: 660/765 becomes 704/765, a $5.75\pp$ gain (paired 95\% interval $[3.95,7.55]\pp$). We use C to illustrate the mechanism and run controls, while reporting the other two searches as well. ``Best'' is a comparison among these three reported candidates, not a separately tuned fourth result.

\begin{figure*}[t]
\centering\includegraphics[width=\textwidth]{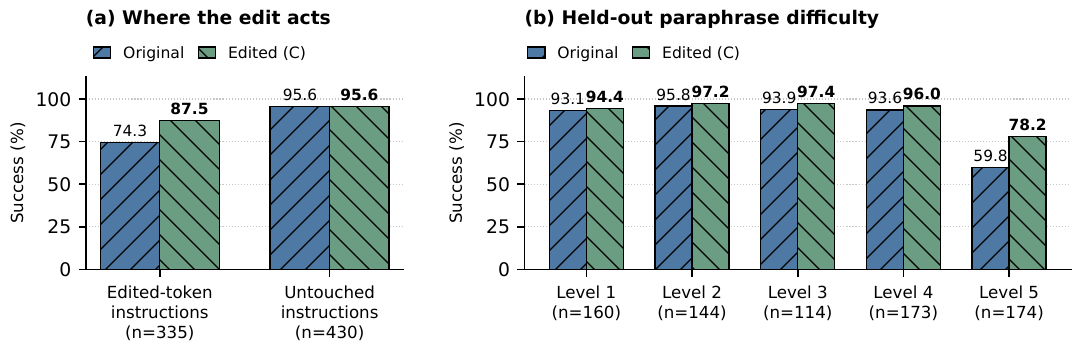}
\caption{\textbf{The repair is local to the instructions it touches.} (a) Run C improves edited-token instructions from 249/335 to 293/335 successes; all 430 untouched instructions retain their outcomes (411 successes). (b) The gain is largest on the hardest paraphrases. All bars are recomputed from episode records and token IDs; these are measured outcomes, not generated imagery.}
\label{fig:scope}
\end{figure*}

Figure~\ref{fig:scope} shows that the largest difficulty-level gain is on level 5, where object and action names are replaced by descriptions: 104/174 becomes 136/174, or 59.8\% to 78.2\%. The selected stove episode in Figure~\ref{fig:teaser} is a level-4 example. Neither example selection nor the difficulty grouping is used to construct the edit.

Level 5 contains 174/765 episodes (22.7\%) but contributes 32 of C's 44 net repairs (72.7\%). The remaining 591 episodes change from 556 to 568 successes, or 94.1\% to 96.1\%. Thus the headline gain is concentrated where descriptions replace familiar task names; it is not a uniform improvement across language difficulty. The edited-token partition is defined separately from difficulty, using the submitted row IDs, and is the stronger test of the proposed lookup mechanism.

\subsection{Only 65 embedding rows}
The embedding table has 257,152 rows of width 2,048. For a selected source row $i$, C writes
\begin{equation}
 E'_i=(1-\alpha_i)E_i+\alpha_i\bar E_{T_i},
 \qquad \bar E_{T_i}=|T_i|^{-1}\!\sum_{j\in T_i}E_j.
 \label{eq:lexical}
\end{equation}
Most destinations are one task-anchor token; ambiguous cues use a centroid. For example, dairy is not forced exclusively toward cheese. Strengths range from 0.5 to 1.0 in the final recipe. Some rows are replaced completely, while others retain part of their original vector. The edit is context-insensitive and does not rewrite the instruction string. It changes 132,768 scalar entries across 65 rows, or 0.00367\% of the exported checkpoint. We independently compare every row of the archived original and edited tensors; all other rows and tensors are unchanged. Appendix~\ref{app:rows} lists every row, destination, and strength.

\begin{table}[t]
\centering\small
\setlength{\tabcolsep}{3.0pt}
\begin{tabular}{lrrrrrr}
\toprule
 & \multicolumn{4}{c}{Edited-token instructions} & \multicolumn{2}{c}{Untouched}\\
\cmidrule(lr){2-5}\cmidrule(lr){6-7}
Run & $n$ & Orig. & Edit & $g/l$ & $n$ & $g/l$\\
\midrule
A & 216 & 67.6 & 78.7 & 32/8 & 549 & 0/0\\
B & 125 & 52.8 & 63.2 & 18/5 & 640 & 0/0\\
C & 335 & 74.3 & \textbf{87.5} & 48/4 & 430 & 0/0\\
\bottomrule
\end{tabular}
\caption{\textbf{Locality holds in all three held-out evaluations.} Rates are percentages. Partitions use exact tokenizer IDs, not substring matching. The affected subset differs across sessions because the edited row sets differ.}
\label{tab:pi05-split}
\end{table}
\paragraph{Locality under fixed inputs.}
Let $S$ be the edited row IDs and $I(c)$ the token IDs read from instruction $c$. On the evaluated path, the policy can be written as
\begin{equation}
 a=F_{\vartheta}\big(V(o),E[I(c)],z\big),
 \label{eq:lookup-policy}
\end{equation}
where $\vartheta$ denotes the unedited parameters, $V(o)$ the visual inputs, and $z$ the fixed flow noise. If $I(c)\cap S=\varnothing$, then $E'[I(c)]=E[I(c)]$. Substitution in Equation~\ref{eq:lookup-policy} gives exactly the same action chunk under deterministic execution. This is a structural locality guarantee: it follows from which rows are read, without assuming that a word is semantically unrelated to the task.

The empirical counterpart is Table~\ref{tab:pi05-split}: every observed outcome change falls inside the edited-token subset. For C, all 430 outside episodes keep their outcomes; within the affected subset, 48 of the original 86 failures are repaired (55.8\%), with a $13.13\pp$ gain and paired 95\% interval $[9.16,17.11]\pp$. The guarantee concerns fixed inputs. It does not imply that future uses of an edited token are safe, or that separately rendered closed-loop simulations are numerically identical.

On the separate 360-episode sample of the other six categories, C changes no outcomes. A and B each repair the same single episode; thus one unique episode, rather than two distinct episodes, differs across their checks. None of those canonical instructions contains an edited token. The record includes small simulator/render variability, so fixed-input policy invariance and closed-loop outcome identity are distinguished in Appendix~\ref{app:language}.

A recorded light-condition stress test checks the distinction directly. Replaying the same saved inputs gives identical original/edited actions and requested residual arrays on all 22 inputs. Separately rendered paired episodes can nevertheless differ: the notebook records initial-image differences of 1--16 channel entries, with a maximum one-level change in an 8-bit channel, while robot state and flow noise match. This explains why a closed-loop discordance on an untouched prompt is not by itself evidence that the weight edit changed the fixed-input policy. We retain such outcomes instead of deleting them as inconvenient observations.

\subsection{What the controls explain}
\begin{table}[t]
\centering\small
\setlength{\tabcolsep}{3.0pt}
\begin{tabular}{lrrr}
\toprule
Same 65 source rows & Success & $g/l$ & Exact $p$\\
\midrule
No edit & 86.3 & -- & --\\
Random direction, matched norm & 85.2 & 7/15 & 0.134\\
Full destination replacement & 90.5 & 43/11 & $1.40\!\times\!10^{-5}$\\
Editor (C) & \textbf{92.0} & 48/4 & $1.31\!\times\!10^{-10}$\\
\bottomrule
\end{tabular}
\caption{\textbf{Controls on the fixed 765-variant population.} Success is in percent; $p$ compares each edit with the original. The full-replacement control uses C's own discovered source--destination pairs, including centroids. Two full-replacement and three random-control runtime errors count as failures here; valid-pair sensitivity checks are in Appendix~\ref{app:language}.}
\label{tab:pi05-controls}
\end{table}
Moving the same rows by matched distances in random directions does not improve success (Table~\ref{tab:pi05-controls}). This control tests the direction, not just the number or norm of changed entries; it is one realized random control, not an average over random seeds. Full replacement toward the editor's destinations explains much of the improvement. C has the higher point estimate and fewer regressions, but its direct comparison with full replacement is not significant at 0.05: 20 wins and nine losses on 763 valid pairs, exact $p=0.0614$.

The control conclusion also survives a change in runtime-error handling. On valid pairs only, full replacement gives 692/763 successes versus an original 659/763; random directions give 652/762 versus 657/762. The corresponding edited rates are 90.7\% and 85.6\%, compared with 90.5\% and 85.2\% when errors count as failures. Neither convention supplies significant evidence that C outperforms the editor-informed replacement control. This sensitivity analysis separates semantic evidence from infrastructure failures without choosing the convention that gives the largest apparent advantage.

The dictionary control is \emph{editor-informed}: it inherits which words to alter and which destinations to use. It does not measure the cost of discovering that lexicon. The main contribution is the diagnosis and construction from rollouts, probes, and weights without a supplied synonym list; the extra benefit of partial and blended moves remains a smaller, statistically unresolved comparison.

\subsection{Causal checks with fixed observations}
Success counts alone do not identify how the edit works. We therefore audit B's recorded replay controls, which keep the image, instruction, flow noise, and action decoder fixed. The comparison separates an embedding's direction from its norm. It also compares predicted actions with those for a matched canonical instruction, rather than using attention mass as a proxy for task understanding.

For seven descriptive inputs, let $a_i$ be the original action chunk, $\widetilde a_i$ the tested chunk, and $a_i^{\rm anchor}$ the chunk for its literal task prompt. We report the median ratio
\begin{equation}
 \rho=\operatorname{median}_i
 \frac{\lVert\widetilde a_i-a_i^{\rm anchor}\rVert_2}
 {\lVert a_i-a_i^{\rm anchor}\rVert_2}.
 \label{eq:anchor-distance}
\end{equation}
We independently recompute these distances from the saved action arrays. The norm-only control leaves $\rho$ near one; direction-only and final edits reduce it to 0.340 and 0.367 (Table~\ref{tab:replay-controls}). Closeness to an anchor is a mechanistic diagnostic, not a substitute for closed-loop task success.

\begin{table}[t]
\centering\small
\setlength{\tabcolsep}{3.7pt}
\begin{tabular}{lrrr}
\toprule
Edit & $\rho$ & Rel. shift & Shift cosine\\
\midrule
Original & 1.000 & 0 & --\\
Norm only & 1.004 & 0.037 & $-0.122$\\
Direction only & 0.340 & 1.039 & 0.983\\
Final B & 0.367 & 1.000 & 1.000\\
\bottomrule
\end{tabular}
\caption{\textbf{Semantic direction carries the replay effect.} $\rho$ is the median over seven descriptive inputs. ``Rel. shift'' and ``Shift cosine'' compare the concatenated action displacement on affected replay inputs with final B. Five untouched control prompts have identical actions under both control edits.}
\label{tab:replay-controls}
\end{table}
For ``activate the cooktop'', distance to the stove anchor changes from 2.409 to 2.418 under the norm control, 0.409 under direction-only editing, and 0.417 under final B. An earlier half-strength candidate gives 0.123, but that selected replay improvement does not make it the best rollout candidate. Repeating original inference reproduces all 16 original chunks exactly. Removing three destination substitutions from final B reduces net repairs from 13 to nine on the same 64 practice episodes. Together these controls support a lexical-direction mechanism while retaining the distinction between a local action change and a validated repair.

\paragraph{Action alignment is not global feature alignment.}
The recorded residual probes also limit the interpretation. For the cooktop input, final B brings the action-expert residual closer to its stove anchor at block indices 0, 9, and 17: distances are 0.282, 0.540, and 0.595 times their original values. Yet the final VLM-language residual is 3.165 times its original distance. These are recorded selected-input diagnostics. They show that successful action alignment need not monotonically reduce every intermediate feature distance. A repair can reconnect a task-relevant action route without reconstructing the literal prompt's complete representation; conversely, an attractive representation metric would not by itself establish a rollout gain.

\subsection{Practice validation beyond one seed}
B also reserves variants before its final validation. We reconstruct the complete 1,533 paired practice trials on 577 non-development variants, including both initial infrastructure errors and their same-seed retries. Reconciled success changes from 1,356 to 1,376 ($25/5$); retaining initial error outcomes gives 1,342 to 1,360 ($34/16$). Both conventions are reported rather than replacing failures silently.

\begin{table}[t]
\centering\small
\setlength{\tabcolsep}{4.1pt}
\begin{tabular}{rrrrrl}
\toprule
Seed & $n$ & Original & Edited & $g/l$ & Exact $p$\\
\midrule
23 & 512 & 451 & 465 & 14/0 & .000122\\
37 & 512 & 458 & 464 & 7/1 & .0703\\
61 & 65 & 58 & 59 & 1/0 & 1.000\\
67 & 63 & 57 & 58 & 1/0 & 1.000\\
73 & 381 & 332 & 330 & 2/4 & .6875\\
\bottomrule
\end{tabular}
\caption{\textbf{B's reserved practice validation by rollout seed.} Counts use reconciled same-seed retries. Variants repeat across seed rows, so the 1,533 trials are not independent environment samples. This is practice validation, separate from the 765-variant held-out experiment.}
\label{tab:practice-seeds}
\end{table}
Table~\ref{tab:practice-seeds} shows why a positive aggregate should not be read as uniform robustness. Seed 23 improves, seed 37 has a smaller positive estimate, and seed 73 loses two net successes. The candidate stays frozen during these checks. The main held-out result uses seed 7 and three independently constructed candidates; this practice evidence adds a test of seed sensitivity, not a claim that every seed confirms the gain. Detailed replay prompts, trial keys, retry handling, and recomputation scripts are provided in the supplement.

\paragraph{What the four regressions look like.}
Two regressions describe a round or rounded food container and an upper storage location; the others describe a salad-dressing container and two espresso makers. The repeated storage description makes the remaining error concrete: lexical cues can be individually familiar while the complete instruction still selects the wrong behavior. These examples do not identify which single row causes each loss; that would require a separate ablation. They instead delimit the current mechanism's scope. The four instructions, token mappings, and outcomes are retained in the supplement, alongside the 48 repairs in the all-pair records. An outcome is not removed because a description is ambiguous or difficult to interpret afterward.

\begin{figure*}[t]
\centering\includegraphics[width=\textwidth]{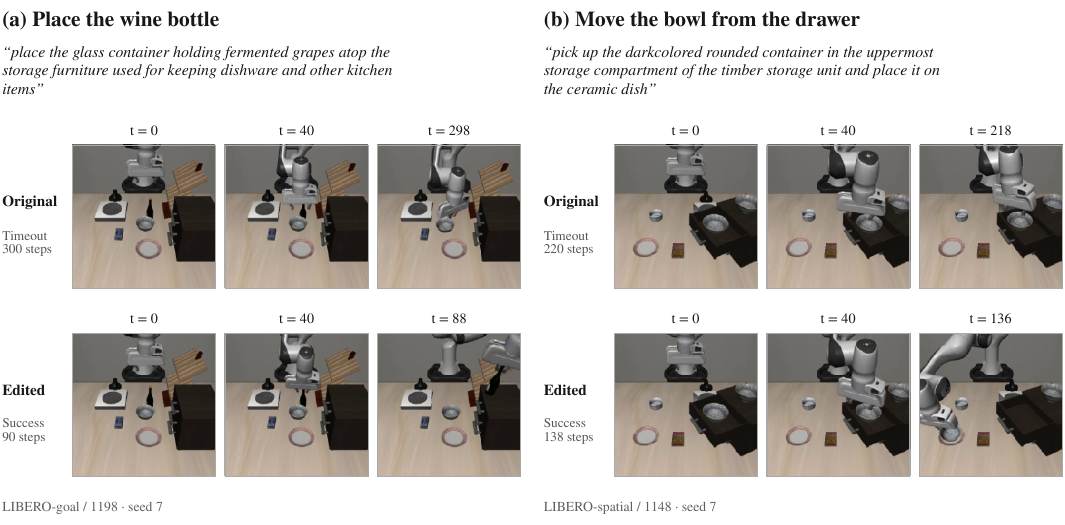}
\caption{\textbf{Additional held-out repairs under the same instructions.} Run C succeeds on (a) wine placement, LIBERO-Goal 1198, and (b) bowl retrieval, LIBERO-Spatial 1148, while the original policy times out. Both use seed 7 and initial state 0. The displayed $t$ values are policy steps; the final recorded frames precede success checks at steps 90 and 138. Video action hashes match the corresponding held-out records. These selected repairs illustrate behavior; Table~\ref{tab:pi05-main} reports all outcomes, including regressions. Expanded sequences and task catalogs are in Appendix~\ref{app:qualitative}.}
\label{fig:additional-examples}
\end{figure*}

\section{When the Editor Declines a Repair}
\label{sec:limits}
\begin{table*}[t]
\centering\small
\setlength{\tabcolsep}{4.5pt}
\renewcommand{\arraystretch}{1.12}
\begin{tabular}{llrrrrrl}
\toprule
Policy & Target / session & Runs & Retained & $n$ & Original & Candidate & Exact $p$\\
\midrule
$\pi_{0.5}$ & Paraphrases A & 1 & 1 & 765 & 86.3 & 89.4 & $1.82\times10^{-4}$\\
 & Paraphrases B & 1 & 1 & 765 & 86.3 & 88.0 & .0106\\
 & Paraphrases C & 1 & 1 & 765 & 86.3 & \textbf{92.0} & $1.31\times10^{-10}$\\
\addlinespace[4pt]
FastWAM & Suffix, first$^{\dagger}$ & 1 & 0 & 700 & 47.7 & 51.3 & .00187\\
 & Suffix, follow-ups A/B & 2 & 0 & -- & -- & -- & --\\
 & Suffix, follow-up C & 1 & 1 & 700 & 48.0 & 48.7 & .542\\
 & Camera$^{\dagger}$ & 1 & 0 & 200 & 33.0 & 36.5 & .230\\
 & Background$^{\dagger}$ & 1 & 0 & 200 & 62.5 & 63.0 & 1.000\\
 & Language$^{\dagger}$ & 1 & 0 & 200 & 60.5 & 59.5 & .791\\
\addlinespace[4pt]
$\pi_{0.5}$ & Failure recovery A/B/C & 3 & 0 & -- & -- & -- & --\\
 & PRO A$^{\dagger}$ & 1 & 0 & 1,484 & 58.5 & 60.0 & .0448\\
 & PRO B/C & 2 & 0 & -- & -- & -- & --\\
\midrule
 & Total sessions & 16 & 4 & & & &\\
\bottomrule
\end{tabular}
\caption{\textbf{Final decisions and held-out outcomes across all 16 sessions.} Success rates are percentages; $n$ counts episode pairs, while ``Runs'' counts editor sessions. $^{\dagger}$Withdrawn candidate evaluated from its archive. Camera/background/language rows use their targeted category; suffix and PRO rows use the stated pooled samples. Dashes mean no candidate result is presented, not zero success. All three paraphrase edits are retained and significant. Of the other 13 sessions, 12 decline their candidates; the sole retained suffix follow-up has $p=.542$.}
\label{tab:limits}
\end{table*}
The second result is the pattern of \emph{declined} interventions. Table~\ref{tab:limits} includes every robot session. Outside the three language repairs, 12 of 13 final decisions revert or decline. The sole retained edit is a FastWAM value-projection candidate whose held-out gain is not significant. We therefore distinguish an editor's ability to influence a pathway, its decision to retain that intervention, and independent evidence of a behavioral repair.

\paragraph{Recovery: a correlate can be a consequence.}
On LIBERO-RECOVER~\cite{liberorecover}, the evaluated policy succeeds on 59.5\% of 440 baseline episodes. Of 178 failures, 141 contain at least six gripper flips. A tempting diagnosis is that gripper looping causes failure. One editor finds instead that failed episodes do not flip more often per step; flips accumulate during long failures. A second mechanistic finding identifies a late visual-to-gripper route: an intervention increases reopening after empty grasps from 5\% to 15\% of the relevant action tokens. Yet its paired practice evaluation changes 213 successes to 208 among 352 episodes. Manipulating the correlate does not establish a repair. All three recovery sessions submit no edit.

\paragraph{New positions and tasks.}
On held-out LIBERO-PRO~\cite{liberopro}, original success is 28.0\% for position changes and 26.3\% for task changes; the policy often completes the scene's original task. Two sessions remove their candidates after validation. A third reduces five outlying SigLIP position-embedding rows. Its 768-episode practice confirmation is 434 versus 433 successes, and it declares withdrawal without executing cleanup. The archived candidate nevertheless changes overall held-out success from 868/1,484 to 890/1,484 ($p=.0448$). The gain concentrates in object appearance; position success falls from 28.0\% to 27.3\%, and the task effect is not significant. We report this nominal aggregate effect but do not promote it to a validated target repair. Subgroup tests are unadjusted.

\paragraph{What the decisions can establish.}
The decision table separates two forms of evidence: validation available to the editor, and held-out assessment available only afterward. A declined candidate can still have a nominal positive held-out estimate, as the first suffix edit and PRO A show. Conversely, a retained candidate can fail independent assessment, as suffix follow-up C shows. Thus the 12/13 decline rate is not a measured accuracy for a failure classifier. It records whether these searches recognize a lack of practice support. The positive language result requires both a retained edit and an independently positive held-out effect.

\paragraph{The editor's own standard.}
The decisions are not inferred retrospectively from held-out scores. Notes and final messages explicitly separate mechanism from efficacy. The LIBERO-PRO editor states, ``The candidate failed validation, and the planned reversion has not been executed.'' A recovery session concludes, ``I did not find a validated recovery improvement within the 50-pass budget.'' The data support a useful tendency to decline unsupported edits in this small sample. They do not establish that an LLM can reliably classify arbitrary failures as repairable or impossible.

\section{A Separate Finding: Instruction Metadata}
\label{sec:suffix}
\begin{figure}[t]
\centering\includegraphics[width=\columnwidth]{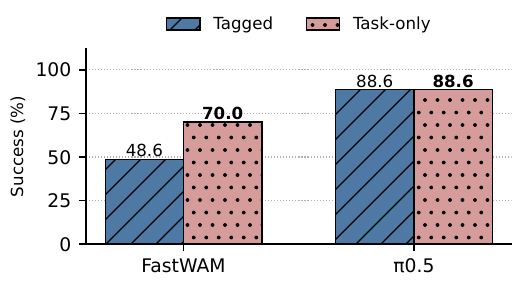}
\caption{\textbf{One instruction format, two different sensitivities.} Each model is tested on the same 210 matched LIBERO-Plus episodes. Keeping reference metadata in the instruction lowers FastWAM success by $21.4\pp$. The evaluated $\pi_{0.5}$ checkpoint has the same aggregate rate in both formats; some individual outcomes differ. This is evidence about these two checkpoints, not all WAMs versus all VLAs.}
\label{fig:suffix}
\end{figure}
LIBERO-Plus's reference instruction can include scene metadata such as \code{view 0 0 100 2 354 initstate 0}. On a fixed 210-episode sample, FastWAM succeeds on 147 episodes with task-only instructions and 102 with tagged instructions: 70.0\% versus 48.6\% (Figure~\ref{fig:suffix}). The evaluated $\pi_{0.5}$ checkpoint scores 186/210, or 88.6\%, with either format. Equal aggregate rates do not mean every paired outcome is identical. Benchmark reporting should therefore state the exact instruction-construction procedure, rather than treating this string as incidental.

The paired counts clarify the aggregate comparison. Removing metadata yields 47 gains and two losses for FastWAM (exact $p=4.36\times10^{-12}$), versus nine gains and nine losses for $\pi_{0.5}$ ($p=1$). Thus unchanged VLA success is an aggregate observation, not invariance of individual trajectories. The two checkpoints also have different overall success and architectures, so the comparison isolates their response to this interface choice rather than proving a universal distinction between WAMs and VLAs.

\paragraph{An interpretable edit that does not replicate.}
The first FastWAM editor constructs a context detector and compiles a negative score contribution into existing text MLPs and Q/K coordinates in both experts. Its archived candidate changes tagged held-out success from 334/700 to 359/700 ($g/l=43/18$, $p=.00187$), but fails a reserved paraphrase preservation check and is requested for withdrawal. The driver does not execute the final cleanup. The architecture and complete gate construction remain in Appendix~\ref{app:gate}, rather than serving as the main success claim.

Three follow-up sessions inherit that notebook and candidate. Two decline their final edits and return original weights. The third keeps a video-expert value-projection edit. On a 700-pair tagged sample it changes success from 48.0\% to 48.7\% ($24/19$, $p=.542$). On a separate 700-pair task-only sample it changes 65.6\% to 66.6\% ($26/19$, $p=.371$). The first-round task-only sample had 420 pairs; these denominators and baselines are not interchangeable.

The notebook evidence points to contextual contamination: adding the suffix shifts representations of task tokens as well as suffix tokens, so suppressing suffix-like features need not restore the original representation. This is a diagnosis supported by probes, not a complete token-level causal explanation. We retain the original $+3.6\pp$ result as an unreplicated, declined candidate. The input-format sensitivity itself is a separate, reproducible comparison and remains relevant to evaluation practice.

\section{Related Work}
\paragraph{Robot foundation policies and world-action models.}
RT-2~\cite{rt2} transfers web representations to robot actions; OpenVLA~\cite{openvla} and Octo~\cite{octo} provide open generalist policy interfaces. The $\pi_0$ family~\cite{pi0,pi05} combines pretrained vision-language representations with flow-matching action generation, building on generative control approaches such as Diffusion Policy~\cite{diffusionpolicy} and flow matching~\cite{flowmatching}. These systems make it possible for familiar physical actions to coexist with fragile task selection. Our checkpoint uses a PaliGemma-style prefix~\cite{paligemma} with SigLIP visual representations~\cite{siglip}; the successful edit changes only the language embedding lookup.

World-action models offer a different route to physical representations. DreamZero~\cite{dreamzero} predicts videos and actions jointly, whereas FastWAM~\cite{fastwam} studies action-only inference from a world-action architecture. E0.5~\cite{e05} uses a 3D foundation model as its native backbone and predicts future geometry together with actions. These approaches broaden the representations through which language may reach control. Our evidence concerns the evaluated VLA and FastWAM checkpoints; whether the same lexical bottlenecks occur in geometry-native policies remains an empirical question.

\paragraph{Editing representations and learned associations.}
ROME, MEMIT, and MEND~\cite{rome,memit,mend} alter learned associations through parameter updates. LEACE~\cite{leace} and inference-time intervention~\cite{iti} expose alternative ways to manipulate representations. Lexical vector alignment has a longer history in word translation~\cite{lexicalalignment}, where mappings between embedding spaces recover word correspondences. Here the editor diagnoses source--destination pairs from a robot's behavior and modifies selected rows inside one existing table. The editor-informed full-replacement control shows that lexical correspondence explains much of the gain; we do not claim a new general embedding-alignment algorithm.

\paragraph{Adaptation, guidance, and evaluation.}
LoRA~\cite{lora} adapts a network with trained low-rank updates; language guidance~\cite{rss} changes how instructions affect inference. Our editing protocol instead fixes deployment instructions and code and excludes gradient fitting. These approaches are relevant alternatives, not baselines we have measured. LIBERO~\cite{libero}, LIBERO-Plus~\cite{liberoplus}, LIBERO-PRO~\cite{liberopro}, and LIBERO-RECOVER~\cite{liberorecover} distinguish nominal success, perturbation robustness, and recovery. We use those distinctions to avoid calling a diagnostic pathway intervention a new manipulation capability.

Weight-space programming~\cite{wsp} supplies the central question: whether an LLM can inspect a network and construct a useful parameter change through tools. Our contribution is the robot study with paired outcomes, sparse edits, mechanism controls, and explicit final retain/revert decisions, including candidates whose held-out results do not support promotion.

\section{Discussion and Limitations}
The title's answer is qualified but affirmative: an LLM can debug a robot policy's lexical instruction grounding directly in its weights. The repair is small, inspectable, and reproducible across three editing sessions. It is also narrow. The edits help instructions containing selected tokens; a word that is a useful synonym in one task can be ambiguous elsewhere. The strongest control already recovers much of the gain using the editor's own discovered lexicon. We have not demonstrated superiority over a lexicon built independently by a human, prompt rewriting, or gradient-based fine-tuning such as LoRA~\cite{lora}.

The obstacle/ability distinction is a productive hypothesis, not a proven boundary. Visual perturbations, recovery states, and task changes yield no retained, independently confirmed repair under the tested budgets. That does not prove no suitable weight edit exists. FastWAM suffix sensitivity is itself obstacle-like, yet the attempted gate is not reproducible. Editability depends on the policy and intervention as well as the task label.

All robot experiments are in simulation and use two released checkpoints. The three primary sessions share a held-out population and seed; their independence is in search, not environmental coverage. The reserved practice-seed checks include a negative seed and are reported in Table~\ref{tab:practice-seeds}. Runtime variation can produce closed-loop discordances even when fixed-input computations are unchanged. The editor consumes observations, activations, and simulator outcomes, so this is not data-free in the broad sense. We report request counts rather than equating a request with an episode or neural forward call; complete API cost accounting is not available.

\paragraph{Artifact availability.}
The accompanying reproducibility package contains the harness and domain code, transcripts for all 16 sessions, paired result records, a sparse 65-row edit with an application script, figure sources, and provenance hashes. Public code and weight-hosting links will be supplied with the artifact release.
\section{Conclusion}
Yes---a language model can repair a robot policy when an existing manipulation skill is blocked by lexical instruction grounding. In our strongest run, modifying only 65 embedding rows increases success from 74.3\% to 87.5\% on affected held-out instructions, while producing zero outcome changes across 430 unaffected instructions. Three independent searches consistently reproduce this improvement, suggesting that the observed gain is not an artifact of a single optimization trajectory. Moreover, twelve of thirteen searches targeting other failures reject their proposed edits, and the sole retained candidate fails to confirm, highlighting the importance of conservative validation when modifying pretrained policies.

These findings establish a concrete and narrowly scoped use for weight-space programming: repairing the connection between familiar linguistic expressions and manipulation behaviors that a robot has already acquired, without retraining the full policy or altering its underlying motor skills. More broadly, they suggest that certain instruction-following failures may arise from localized representational mismatches rather than deficiencies in the robot's action capabilities. Nevertheless, our results do not establish broader language generalization, transfer to physical robots, or the ability to synthesize missing manipulation skills. Addressing these limitations remains an important direction for future work.

\clearpage
\begingroup\small
\balance
\bibliographystyle{IEEEtranN}
\bibliography{references}
\endgroup
\clearpage
\nobalance
\onecolumn
\setcounter{page}{1}\renewcommand{\thepage}{S\arabic{page}}

\vspace{10pt}
\noindent\textbf{Contents and purpose.}
\begin{center}
\small\renewcommand{\arraystretch}{1.2}\setlength{\tabcolsep}{5pt}
\begin{tabular}{p{.28in}p{2.22in}p{4.0in}}
\toprule
 & Material & Evidence added to the main paper\\
\midrule
\hyperref[app:language]{A} & Sampling, statistics, and replay audit & Defines paired intervals and exact tests, discloses the earlier 200-episode sample, separates error/retry conventions, and gives full replay prompts behind the mechanism controls.\\
\hyperref[app:gate]{B} & Archived FastWAM gate construction & Specifies the context direction, reused MLP coordinates, normalized Q/K score contribution, and parameter count. This is the withdrawn suffix candidate, not the main successful repair.\\
\hyperref[app:outcomes]{C} & Detailed FastWAM and PRO outcomes & Gives category counts, original baselines, sample sizes, and follow-up results, including the nominal aggregate gain of a declined PRO candidate.\\
\hyperref[app:record]{D} & Execution and artifact provenance & Distinguishes a final retain/revert decision from files left on disk; documents cleanup behavior, tool budgets, video hashes, and the release contents.\\
\hyperref[app:qwen]{E} & Qwen reproduction protocol & Separates matched 100-problem comparisons from the full edited MATH-500 evaluation and reports token-limit differences from the original study.\\
\hyperref[app:rows]{F} & Complete embedding-row inventory & Lists all 65 token IDs, source pieces, destinations, and interpolation strengths. Exact row values and sparse application code accompany the machine-readable recipe.\\
\hyperref[app:qualitative]{G} & Additional visual examples & Expands four verified held-out trajectory pairs, shows 22 practice task images and four FastWAM input-format pairs, and preserves all four language regressions.\\
\bottomrule
\end{tabular}
\end{center}

\noindent\textbf{Reading the evidence.}
A practice trial is feedback available during editing; a held-out trial is evaluated after candidate selection and is unavailable to the editor. A replay holds observations and flow noise fixed, whereas a rollout includes simulator dynamics and rendering. A retained edit follows the editor's final decision; an archived edit may have been explicitly declined. These distinctions are used consistently in the section headings, captions, and source manifests. References follow the numbering in the main paper.

\vspace{9pt}
\noindent\textbf{Recomputing the numbers.}
The evidence download includes \code{audit\_new\_results.py} for held-out language, controls, input-format sensitivity, and PRO outcomes. The companion \code{audit\_extended.py} recomputes the added practice-seed and replay analyses from archived per-episode JSONL and action arrays. The latter uses NumPy; neither script loads a policy or performs training. The paper-source download contains portable figure scripts and recorded images, while separate downloads provide the sparse edit and all 16 editor transcripts.

\clearpage
\twocolumn
\appendix
\setcounter{figure}{0}\renewcommand{\thefigure}{S\arabic{figure}}
\setcounter{table}{0}\renewcommand{\thetable}{S\arabic{table}}
\setcounter{equation}{0}\renewcommand{\theequation}{S\arabic{equation}}
\section{Additional Language Evaluations and Statistics}
\label{app:language}
For paired binary outcomes, write $d_i=y_i'-y_i\in\{-1,0,1\}$ and $\widehat\Delta=(g-l)/n$. The displayed normal-approximation interval is
\begin{equation}
 100\left[\widehat\Delta\ \pm\ 1.96\sqrt{\frac{(g+l)/n-\widehat\Delta^2}{n}}\right].
 \label{eq:paired}
\end{equation}
We use the exact two-sided binomial test on $g+l$ discordances for significance. A normal interval can exclude zero while the discrete exact test does not; exact $p$ values govern statements of significance. An interval of zero width for an unchanged sample describes its empirical discordances, not proof that the population effect is zero.

\paragraph{The earlier 200-episode sample.}
The initial evaluation uses 200 held-out language episodes and the same 360 other-category episodes. Table~\ref{tab:initial-language} reports every language candidate. The later 765-episode evaluation enumerates all Language Instructions variants in the existing held-out split; no candidate was changed between these evaluations. The baseline contains 105 failures in the full population versus 19 in the smaller sample. These evaluations are disclosed separately, rather than pooled or described as independently replicated datasets.
\begin{table}[ht]
\centering\small
\begin{tabular}{lrrrr}
\toprule
Run & Original & Edited & $g/l$ & Exact $p$\\
\midrule
A & 181/200 & 182/200 & 3/2 & 1.000\\
B & 181/200 & 184/200 & 3/0 & 0.250\\
C & 181/200 & 185/200 & 6/2 & 0.289\\
\bottomrule
\end{tabular}
\caption{Initial held-out language sample; none of the changes is significant.}
\label{tab:initial-language}
\end{table}

\paragraph{Extended practice validation audit.}
The main paper reports B's five rollout-seed groups, including the negative seed-73 result. We independently reconstruct every paired trial from requests 18--21, 33--37, and 48--50; request 24 supplies same-seed retries for the earlier infrastructure errors. A trial key is suite, variant, rollout seed, and initial state. Retry rows replace the affected original/edited entries at the same key rather than adding trials. Non-language camera retries in request 50 do not enter the language total. The resulting 1,533 pairs on 577 variants give 1,356 versus 1,376 successes ($25/5$); initial error outcomes give 1,342 versus 1,360 ($34/16$). The independent recomputation agrees with the editor's final transcript. Repeated variants make the pooled episode-level $p$ descriptive rather than a test over 1,533 independent task instances.

\paragraph{Fixed-input causal probes.}
For the final B candidate, replay requests 23, 38, and 39 compare full, norm-only, and direction-only embedding values on the same recorded inputs. The norm/direction split is applied to the selected rows, with the rest of the model unchanged. The concatenated displacement statistics use only inputs affected by the final edit; the seven prompt--anchor pairs below define the action-distance summary. Five literal/untouched controls have identical actions under both controls, and all 16 original chunks reproduce exactly. Distances are Euclidean norms in the backend's executed action space, without additional fitting or normalization except the ratios defined in the main paper.

\begin{table*}[t]
\centering\small\setlength{\tabcolsep}{4pt}
\renewcommand{\arraystretch}{1.1}
\begin{tabular}{p{2.8in}p{1.45in}rrrr}
\toprule
Descriptive prompt & Literal anchor & Orig. & Norm & Dir. & Final\\
\midrule
activate the cooktop & id like you to heat up the stove & 2.409 & 2.418 & 0.409 & 0.417\\
activate the cooking appliance for heating food & id like you to heat up the stove & 1.606 & 1.671 & 0.203 & 0.234\\
activate the cooking appliance & id like you to heat up the stove & 0.706 & 0.748 & 0.350 & 0.337\\
place the container of fermented grape liquid on the elevated storage unit & for safety reasons make sure to place the wine bottle on top of the cabinet & 1.121 & 1.057 & 0.175 & 0.153\\
deposit the spreadable dairy product into the mixing container & ensure the cream cheese is in the bowl & 2.298 & 2.301 & 2.135 & 2.151\\
would you mind putting the cream cheese in the mixing bowl for me & ensure the cream cheese is in the bowl & 0.245 & 0.243 & 0.194 & 0.195\\
place the rounded container used for holding food on the higher storage unit often found in kitchens & gently place the bowl on the cabinet making sure its centered and secure & 0.519 & 0.577 & 0.176 & 0.190\\
\bottomrule
\end{tabular}
\caption{\textbf{All seven prompt--anchor replay comparisons behind the causal summary.} Each number is distance to the original policy's matched literal-anchor action chunk. These are controlled practice inputs, not held-out outcome samples.}
\label{tab:replay-prompts}
\end{table*}

\paragraph{The four held-out regressions.}
The all-variant result includes every changed outcome. Table~\ref{tab:regression-prompts} preserves all four losses, rather than letting the positive examples define the repair's scope. The exact instructions and difficulty levels support the main text's discussion of ambiguous descriptions; per-token intervention ablations were not performed for these failures.

\begin{table*}[t]
\centering\small
\setlength{\tabcolsep}{5pt}
\renewcommand{\arraystretch}{1.15}
\begin{tabular}{lp{4.7in}r}
\toprule
Suite / variant & Exact held-out instruction & Level\\
\midrule
Goal / 1265 & place the rounded container used for holding food on the upper storage unit typically found in kitchens & 4\\
Object / 1111 & grasp the container for enhancing the flavor of greens and relocate it to the woven holder for carrying items & 2\\
Goal / 1280 & place the round container used for holding food onto the higher storage area for kitchen items & 1\\
10 / 1414 & position both italian espresso makers on the cooking surface & 5\\
\bottomrule
\end{tabular}
\caption{\textbf{All four regressions of run C.} Original policy success becomes edited-policy failure in each row (seed 7, initial state 0). Together with 48 repairs, these yield the reported net gain of 44/765.}
\label{tab:regression-prompts}
\end{table*}

\paragraph{Runtime errors in controls.}
The primary table counts two full-replacement and three random-control errors as failures, matching the archived evaluator's aggregate convention. With invalid pairs excluded, full replacement gives 692/763 versus 659/763 originally ($43/10$, $p=5.55\times10^{-6}$). Random direction gives 652/762 versus 657/762 ($7/12$, $p=.359$). The editor has no runtime errors. Its direct comparison with full replacement uses their 763 common valid pairs: 703 versus 692 successes, $20/9$, $p=.0614$. Thus error handling does not alter the qualitative conclusion.

\paragraph{Other categories and fixed-input invariance.}
The 360 non-language prompts contain no edited token IDs in any language run. C has identical success outcomes throughout. A and B each change the same episode (LIBERO-Object variant 2387, seed 7) from failure to success: 302/360 becomes 303/360. This is one unique changed episode across runs, with two run-specific flips. Since the edited rows are never read, fixed-input predictions are unchanged; small rendering or simulator differences can still change a closed-loop episode. We do not attribute this isolated gain to the embedding edits.

\paragraph{Tokenization and checkpoint details.}
The input is stripped, underscores and newlines are replaced with spaces, and the checkpoint's SentencePiece tokenizer adds a beginning-of-sequence token and terminal newline. Instructions are padded to 200 tokens. Partitions in the main paper use token IDs from this exact procedure. The evaluated \code{pi05\_libero} configuration sets \code{discrete\_state\_input=False}; therefore robot state is not serialized into the language prefix for these experiments. Action horizon, executed prefix, and flow steps are ten, five, and ten, respectively. Tokenization, model execution, action scaling, and checkpoint selection are fixed for all comparisons.

\paragraph{Counting parameters.}
The exported checkpoint has 3,616,757,520 scalar parameters. Exact elementwise comparison finds 36,764 changed entries in A, 40,822 in B, and 132,768 in C (0.00102\%, 0.00113\%, and 0.00367\%). Row counts include entries whose individual coefficients happen to remain equal after rounding. Consequently, 65 rows is not the same as 65 scalar weights or 65 lines of Python.

\section{FastWAM: Archived Gate Construction}
\label{app:gate}
The following documents the first suffix candidate, which the editor requested to withdraw. Its positive held-out estimate is not presented as a completed or replicated repair. We preserve its mechanism because it makes the scope of a weight-space intervention concrete.
\begin{figure*}[t]
\centering\includegraphics[width=\textwidth]{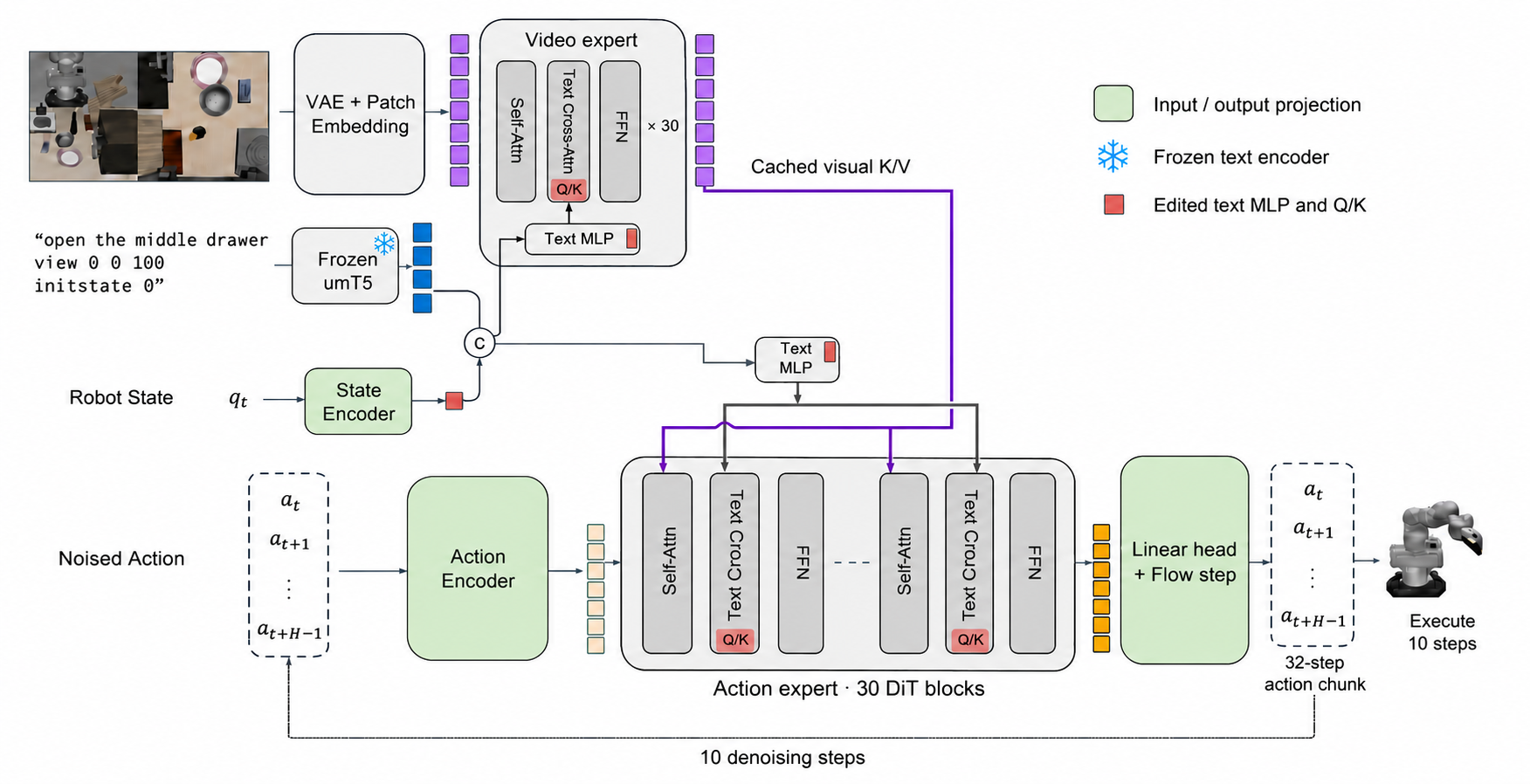}
\caption{\textbf{The FastWAM editing and execution paths.} The previously selected architecture illustration is retained here for the archived suffix investigation. Orange marks rewritten MLP and Q/K weights. FastWAM's execution graph and tensor shapes remain fixed. Camera and action insets are schematic; the diagram does not describe the successful $\pi_{0.5}$ embedding edit.}
\label{fig:fastwam-architecture}
\end{figure*}
\subsection{Problem and policy structure}
Let $c$ denote a task instruction, $s$ a scene-variant suffix, $o$ the camera observations, and $p$ the robot state. We compare a policy conditioned on the task-only string $c$ with the same policy conditioned on $c\oplus s$. The editing objective is to improve closed-loop success under $c\oplus s$ while preserving useful behavior under $c$ and paraphrases of $c$. Only the parameter values $\theta$ may change.

FastWAM has a video expert with hidden width 3072 and an action expert with hidden width 1024. Each has 30 blocks and 24 attention heads of width 128. In the evaluated action-only inference, the video expert prefills layer-wise K/V caches once per policy query. At each denoising step, action queries attend to the cached video K/V and the current action K/V. Each expert separately reads text through cross-attention. A frozen umT5-XXL encodes the instruction, and an encoded proprioceptive token is appended to the context. Nuisance text can therefore affect actions through both the action expert and the video-conditioned route.

\subsection{Constructing a context detector}
Astra begins with 24 raw/task-only practice pairs. The original policy succeeds on 13/24 tagged episodes and 20/24 task-only episodes. Initial images, proprioception, and seeds match. The available attention-map interface exposes action-to-video attention; localization within text conditioning therefore proceeds through weight interventions and replay measurements.

\begin{figure}[ht]
\centering
\includegraphics[width=\columnwidth]{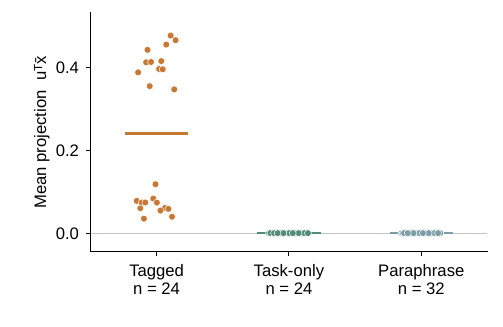}
\caption{\textbf{Observed context means used to construct $u$.} Each point is one saved context mean projected onto the chosen direction; bars show group means. Task-only and paraphrase means are near zero by construction. These are construction diagnostics, not held-out evidence or token-level nonlinear gate responses.}
\label{fig:context-projections}
\end{figure}

After several projection and head-ablation candidates, Astra uses temporary tensor rewrites to read 4096-dimensional context means through the forward tool. It collects 24 raw/task-only pairs and 32 separate paraphrase examples, then removes the instrumentation before repair testing. Let $d_i=\bar{x}^{\rm raw}_i-\bar{x}^{\rm clean}_i$ be a paired context-mean difference. Let $Q$ contain the thin-SVD orthonormal columns of the matrix formed from observed clean means, paraphrase means, and the proprioceptive encoder's weight/bias span. The selected direction is
\begin{equation}
 u=\frac{(I-QQ^\top)\bar d}{\lVert(I-QQ^\top)\bar d\rVert_2},
 \qquad \bar d=\frac1{24}\sum_{i=1}^{24}d_i,
 \label{eq:direction}
\end{equation}
with a further numerical projection against the proprioceptive span and sign chosen toward tagged contexts. Figure~\ref{fig:context-projections} plots the resulting projections of the recorded means. The construction protects the measured mean subspace; it does not guarantee invariance of every clean token or action.

The detector is evaluated on individual context tokens through an existing text MLP:
\begin{equation}
 g(x_t)=\eta^{-1}\operatorname{GELU}\!\left(\eta(u^\top x_t-\tau)\right),
 \label{eq:detector}
\end{equation}
where the frozen candidate uses $\eta=32$ and $\tau=1$. Astra explores a finite set of directions, gains, and thresholds using practice probes. This is a context-feature detector rather than an explicit parser of suffix strings.

\subsection{Compiling the gate into existing weights}

\begin{figure*}[t]
\centering\includegraphics[width=\textwidth]{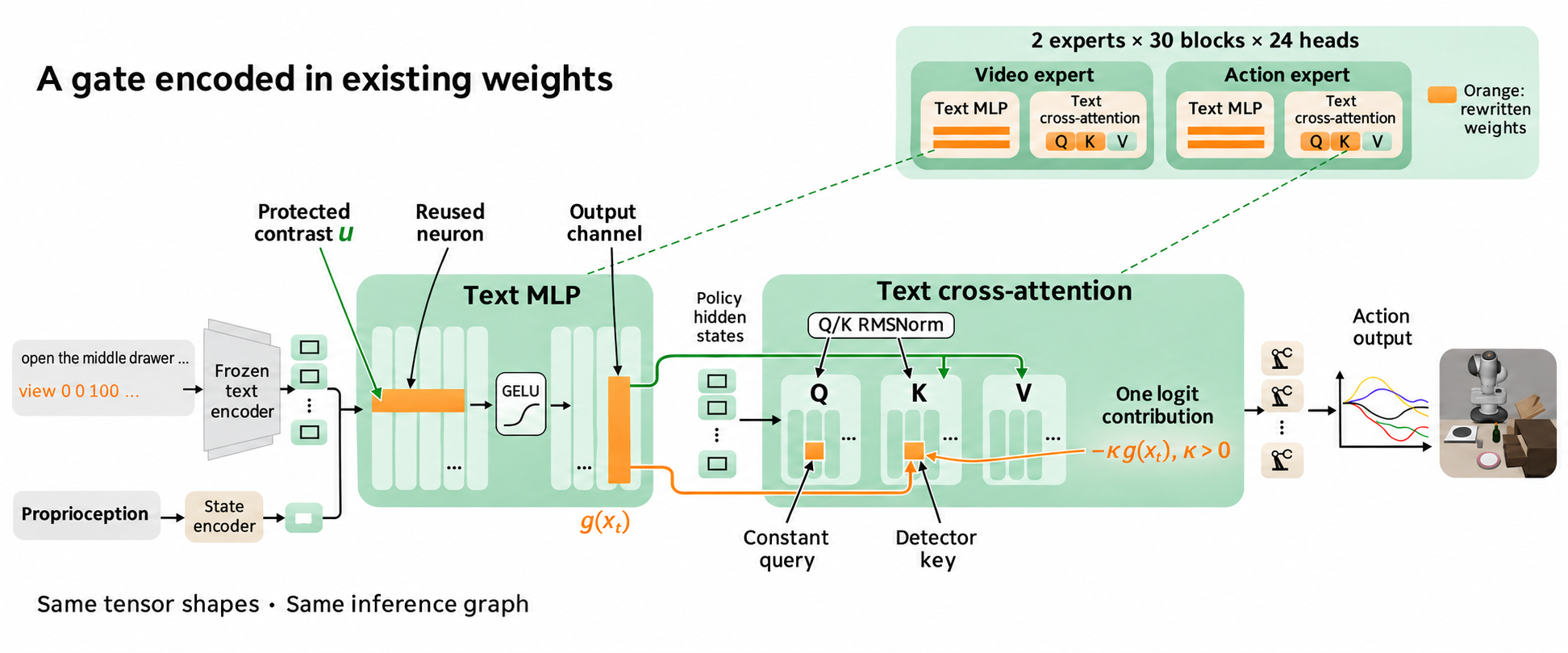}
\caption{\textbf{Gate construction in existing FastWAM coordinates.} The protected context direction controls an existing MLP neuron and output channel, which feeds a reassigned Q/K coordinate. The negative term describes one contribution to the normalized attention score, not a measured attention map or an exact additive mask on the original logits. Schematic generated illustration.}
\label{fig:gate}
\end{figure*}

Figure~\ref{fig:gate} and Algorithm~\ref{alg:gate} specify the candidate. In each expert, one low-energy hidden neuron $j$ and output channel $d$ are reused. The first text-MLP layer writes $\eta u^\top x_t-\eta\tau$ into neuron $j$. The second layer retains only the connection $1/\eta$ from $j$ to output $d$, producing $g(x_t)$ there. Its other contributions to channel $d$, and neuron $j$'s contributions to other channels, are zeroed.

For head $h$ in block $\ell$, the edit chooses a low-energy coordinate $r$ within the head. Replacing the query row with zero and its bias with $c_\ell>0$ gives $q'_r=c_\ell$. Replacing the key row by $-\lambda e_d^\top$ with zero bias gives $k'_r=-\lambda g(x_t)$, where $\lambda=64$. The query RMS-normalization gain is set to $\gamma_{q,r}=\alpha/\gamma_{k,r}$ with $\alpha=6$.

Writing $\sigma_q,\sigma_k>0$ for the RMS normalization denominators, this coordinate contributes
\begin{equation}
 \frac{\bar q'_r\bar k'_r}{\sqrt{128}}
 =-\frac{\alpha\lambda c_\ell g(x_t)}{\sqrt{128}\,\sigma_q\sigma_k}
 \label{eq:score}
\end{equation}
to the modified attention logit. The contribution is negative when $g(x_t)>0$. This equation describes the reassigned coordinate; the overall edit also replaces an old contribution and changes normalization, so it is not an exact additive mask on the original attention scores.

The gate is installed in both experts, across all 30 text-cross-attention blocks and all 24 heads. Four text-MLP tensors and five query/key-related tensors per block are edited in each expert, for $2(4+30\times5)=308$ tensors. The text encoder, model graph, instruction strings, and inference settings are unchanged. Tensor count does not measure the fraction of scalar parameters changed.

\begin{algorithm}[t]
\caption{Compile a context gate into policy weights}
\label{alg:gate}
\DontPrintSemicolon
\KwIn{Original weights $\theta_0$; direction $u$; $\tau=1$, $\eta=32$, $\lambda=64$, $\alpha=6$}
$\theta'\leftarrow\theta_0$\;
\ForEach{expert $e\in\{\mathrm{action},\mathrm{video}\}$}{
Select low-energy text-MLP neuron $j$ and output $d$\;
$W_{0,j:}\leftarrow\eta u^\top$; $b_{0,j}\leftarrow-\eta\tau$\;
$W_{2,:j}\leftarrow0$; $W_{2,d:}\leftarrow0$; $W_{2,dj}\leftarrow1/\eta$; $b_{2,d}\leftarrow0$\;
\ForEach{block $\ell$ and head $h$}{
Select a low-energy coordinate $r$ in head $h$\;
$W_{q,r:}\leftarrow0$; $b_{q,r}\leftarrow c_\ell$\;
$W_{k,r:}\leftarrow-\lambda e_d^\top$; $b_{k,r}\leftarrow0$\;
$\gamma_{q,r}\leftarrow\alpha/\gamma_{k,r}$\;
}
}
\KwRet{Replacement tensors in $\theta'$}
\end{algorithm}
\paragraph{Selecting a coordinate to reuse.}
For each expert, the text-MLP hidden neuron minimizes the product of its incoming-row and outgoing-column weight norms. The output channel minimizes its second-layer row energy plus squared bias. For a cross-attention coordinate $r$, the editor forms a query-energy proxy including the preceding LayerNorm affine transform:
\begin{align}
 E_q(r)&=\lVert W_{q,r:}\operatorname{diag}(\gamma_3)\rVert_2^2
 +(W_{q,r:}\beta_3+b_{q,r})^2,\\
 E_k(r)&=\lVert W_{k,r:}\rVert_2^2+b_{k,r}^2.
\end{align}
The selected coordinate minimizes $\sqrt{E_q(r)E_k(r)}\,|\gamma_{q,r}\gamma_{k,r}|$ within each head. The constant $c_\ell$ is the square root of the mean query-energy proxy in that block. The selection is based on original weights, not a learned importance estimator.

\paragraph{Mean protection and detector selection.}
The protected matrix contains 24 task-only context means, 32 paraphrase means, and the nine columns formed from the proprioceptive encoder's weights and bias. Thin SVD supplies $Q$ in Equation~\ref{eq:direction}; the implementation also removes the proprioceptive span numerically before unit normalization. The detector is chosen from eight directions, 32 thresholds, and two gains tested through a temporary readout. Orthogonality protects the detector's response on the represented span; reassigning existing MLP and attention coordinates can still alter the policy's behavior on clean inputs.

\begin{table}[ht]
\centering\scriptsize
\setlength{\tabcolsep}{2.6pt}
\begin{tabular}{lrrrr}
\toprule
Intervention & Request & $R$ & Clean drift & Para. drift \\
\midrule
Rank-4 projection & 22 & 1.0047 & 0.00316 & 0.00254 \\
Action-only gate, $\tau=.5$ & 24 & 0.9949 & 0.00115 & 0.00111 \\
Both-expert gate, $\tau=.5$ & 25 & 0.9684 & 0.00131 & 0.00137 \\
Rank-24 oblique & 26 & 0.9683 & 0.00289 & 0.00384 \\
Prefix-output projector & 31 & 1.1037 & 0.02956 & 0.00960 \\
Four-detector gate & 32 & 1.0224 & 0.00136 & 0.00147 \\
Both experts, $\tau=1$ & 33 & 0.9657 & 0.00131 & 0.00137 \\
\bottomrule
\end{tabular}
\caption{Practice replay diagnostics transcribed from executed tool outputs and the editor's notebook. $R$ is the raw-to-clean-reference action-gap ratio used in the practice replay probe. Drift is the corresponding original/edited action discrepancy on saved task-only or paraphrase inputs, in the tool's normalized action space. Candidates are sequential diagnostics, not randomized rollout ablations.}
\label{tab:allprobes}
\end{table}

\paragraph{Attention-score interpretation.}
The cross-attention implementation normalizes the full projected query and key with learned RMSNorm gains before splitting into heads. The selected coordinate's gain product is $\alpha$, giving Equation~\ref{eq:score}. Since the normalization denominators depend on the modified vectors, editing one coordinate affects more than one numerical term. A positive detector gives a negative reassigned-coordinate contribution, while full logit differences need not equal that term. No token-level cross-text attention maps are available in this run.

\section{FastWAM and LIBERO-PRO Outcome Details}
\label{app:outcomes}
\begin{table}[ht]
\centering\small
\begin{tabular}{lrrrr}
\toprule
Candidate / format & $n$ & Original & Edit & $g/l$\\
\midrule
First / tagged & 700 & 334 & 359 & 43/18\\
First / task-only & 420 & 291 & 293 & 14/12\\
Follow-up C / tagged & 700 & 336 & 341 & 24/19\\
Follow-up C / task-only & 700 & 459 & 466 & 26/19\\
\bottomrule
\end{tabular}
\caption{Different sample sizes and original counts across suffix evaluations. First-candidate $p$ values: .00187/.845; follow-up C: .542/.371. Follow-ups A and B return the original weights.}
\end{table}
\begin{table*}[t]
\centering\small
\setlength{\tabcolsep}{4.2pt}
\begin{tabular}{lrrrrrrrr}
\toprule
 & \multicolumn{4}{c}{Tagged held-out ($n=100$ per category)} & \multicolumn{4}{c}{Task-only held-out ($n=60$ per category)} \\
\cmidrule(lr){2-5}\cmidrule(lr){6-9}
Category & Orig. & Edit & $\Delta\pp$ [95\% CI] & $g/l$ & Orig. & Edit & $\Delta\pp$ [95\% CI] & $g/l$ \\
\midrule
Background & 45.0 & 43.0 & $-2.0\ [-6.8,2.8]$ & $2/4$ & 56.7 & 58.3 & $+1.7\ [-1.6,4.9]$ & $1/0$ \\
Camera & 7.0 & 13.0 & $+6.0\ [0.6,11.4]$ & $7/1$ & 30.0 & 28.3 & $-1.7\ [-10.3,7.0]$ & $3/4$ \\
Language & 65.0 & 68.0 & $+3.0\ [-0.3,6.3]$ & $3/0$ & 85.0 & 80.0 & $-5.0\ [-10.5,0.5]$ & $0/3$ \\
Lighting & 75.0 & 73.0 & $-2.0\ [-6.8,2.8]$ & $2/4$ & 95.0 & 95.0 & $+0.0\ [-6.5,6.5]$ & $2/2$ \\
Layout & 58.0 & 60.0 & $+2.0\ [-1.9,5.9]$ & $3/1$ & 83.3 & 81.7 & $-1.7\ [-4.9,1.6]$ & $0/1$ \\
Robot state & 45.0 & 52.0 & $+7.0\ [0.6,13.4]$ & $9/2$ & 68.3 & 73.3 & $+5.0\ [-0.5,10.5]$ & $3/0$ \\
Noise & 39.0 & 50.0 & $+11.0\ [1.9,20.1]$ & $17/6$ & 66.7 & 71.7 & $+5.0\ [-3.5,13.5]$ & $5/2$ \\
\midrule
Overall & 47.7 & 51.3 & $+3.6\ [1.4,5.7]$ & $43/18$ & 69.3 & 69.8 & $+0.5\ [-1.9,2.9]$ & $14/12$ \\
\bottomrule
\end{tabular}
\caption{\textbf{Category outcomes for the archived first suffix candidate.} Original and candidate success rates are percentages. Paired difference intervals follow Equation~\ref{eq:paired}; $g/l$ counts improvements/regressions. Category comparisons are descriptive, without multiplicity adjustment. The two instruction conditions use distinct samples. Overall sample sizes are 700 and 420.}
\label{tab:category}
\end{table*}

\paragraph{Other round-one investigations.}
Separate 40-request camera, background, and language investigations were evaluated on an 840-episode task-only sample. Aggregate success changes from 59.4\% to 61.4\%, 59.6\%, and 59.8\%, respectively, with exact paired $p$ values 0.0604, 0.754, and 0.784. These exploratory artifacts have the same finalization caveat as the suffix run. They do not establish three additional successful repairs.

\begin{table}[ht]
\centering\small
\begin{tabular}{lrrrr}
\toprule
PRO category & $n$ & Original & Edit & $g/l$\\
\midrule
All Perturbed & 1484 & 868 & 890 & 66/44\\
Object & 297 & 253 & 266 & 20/7\\
Position & 300 & 84 & 82 & 11/13\\
Semantic & 300 & 288 & 292 & 6/2\\
Task & 300 & 79 & 85 & 11/5\\
Environment & 287 & 164 & 165 & 18/17\\
Original & 200 & 192 & 193 & 5/4\\
\bottomrule
\end{tabular}
\caption{Archived LIBERO-PRO A, common valid pairs only. Sixteen of 1,700 requested original/perturbed pairs are invalid and excluded; 1,484 perturbed and 200 original pairs remain. The targeted position and task groups do not show significant gains. The 1,484-pair aggregate excludes the 200 original episodes.}
\end{table}

\section{Execution Records and Reproducibility}
\label{app:record}
\paragraph{Final intent versus files.}
We count a candidate as retained only if the editor's final decision retains it. In the first FastWAM suffix run, the last executed tool call is turn 111. Turn 112 emits Python that removes all 308 replacement tensors and restores the original checkpoint, but the old driver does not execute that block. The current finalization branch permits up to three Python cleanup executions after the forward budget is exhausted. The first camera, background, and language searches have the same archival caveat.

LIBERO-PRO A is a distinct case: the editor declares withdrawal but emits no cleanup code. Its final message states, ``The candidate failed validation, and the planned reversion has not been executed.'' We therefore report its archived held-out result without calling it a retained repair or claiming that the driver ran an unissued reversion. The other two PRO sessions and all three Recovery sessions retain no edit. The latter decisions express failure to find a validated edit within the budget, not an impossibility result.

\paragraph{Follow-up and negative checks.}
The three suffix follow-ups each have 50 forward requests and inherit prior notes and a candidate; they are not independent rediscoveries. The three language searches are separate 50-request sessions. Recovery C's paired 352-episode practice check changes 213 successes to 208 ($17/22$); increased gripper reopening in a replay probe does not reverse this negative success result. PRO A's reserved practice confirmation changes 434/768 to 433/768. Both are reported rather than selecting only discovery batches. The 440-episode Recovery scan and the 141-of-178 gripper-flip count are diagnostic summaries from the supplied study record; they are not a new held-out comparison in this paper.

\paragraph{Tool budget.}
A CPU execution has a 900-second/96-GB limit. FastWAM requests support at most 256 total rollout episodes, or 128 per model for paired evaluation, and replay probes up to 512 inputs. Backend limits differ and are specified with the released domain configurations. Requests can contain many steps and episodes, so a 50-request budget is not 50 neural forward calls. Conversation handoffs preserve the task and saved notes. No evaluation in this revision performs new policy training or gradient computation.

\paragraph{Trajectory provenance.}
Figure~\ref{fig:teaser} uses saved frames for LIBERO-Goal variant 1421, seed 7, initial-state index 0. The instruction is ``activate the cooking appliances heat source''. The original trace terminates at the 300-step cap; the candidate succeeds at step 78. Their last recorded frames are steps 298 and 76, respectively; recording occurs before the terminal check. The action-log hashes are \code{4a6566f0af5c2ae1} and \code{5d1de45ce39b18fa}. The paired videos use the archived H.264 encodings. A second example (variant 1198) is available on the project page. These are selected improved episodes, not substitutes for all-pair outcome tables.

\paragraph{Release contents.}
Local downloads contain the source harness and domains, all 16 editor transcripts and notes, per-episode outcomes including runtime errors, exact recomputation code, a sparse 65-row payload and application script, tensor/record hashes, editable numerical-figure sources, and the paper's BibTeX file. The sparse payload stores original and edited rows, avoiding redistribution of a full base checkpoint. Figure~\ref{fig:workflow} is a generated methods illustration; actual rollout frames and numerical figures are never generated. Public code and weight-hosting links will accompany the artifact release.

\section{Qwen Reproduction Details}
\label{app:qwen}
The released Qwen3.5-0.8B \code{p50c} edit is reconstructed and its tensor hashes checked against the source record. With a 32,768-token generation limit, one response per MATH-500 problem gives 247/500 correct (49.4\%), compared with 48.1\% in the original four-sample report. Thirty percent of local responses reach the token cap, compared with 33.1\% reported. On a fixed matched 100-problem subset, the local original checkpoint gives 9/100 and the edited checkpoint 57/100; the original report's subset baseline is 8.8\%. The local original checkpoint was not evaluated on all 500 problems, so 49.4\% versus 9.0\% is not a matched full-benchmark improvement. The reproduction verifies a released language-model edit, while the robot candidates are newly constructed by the editor.

\section{Complete 65-Row Edit Inventory}
\label{app:rows}
The row recipe below is extracted from the archived C override and tokenizer, then verified by a full comparison of the 257,152-row original and edited tables. An underscore denotes SentencePiece's word-boundary marker, not an extra character inserted into an instruction. Destination lists denote an arithmetic centroid; $\alpha$ is the interpolation strength in Equation~\ref{eq:lexical}. The companion JSON retains exact token IDs and cosine values. Table~\ref{tab:alignment} reports selected measured alignments.
\begin{table*}[t]
\centering\small
\setlength{\tabcolsep}{5pt}
\begin{tabular}{lrr@{\hspace{25pt}}lrr}
\toprule
Token $\rightarrow$ destination & Before & After & Token $\rightarrow$ destination & Before & After\\
\midrule
cooker $\rightarrow$ stove & 0.927 & 1.000 & activate $\rightarrow$ turn & 0.909 & 0.992\\
cooking $\rightarrow$ stove & 0.933 & 0.996 & ignite $\rightarrow$ turn & 0.907 & 0.987\\
basin $\rightarrow$ bowl & 0.929 & 1.000 & relocate $\rightarrow$ put & 0.885 & 0.985\\
fermented $\rightarrow$ wine & 0.937 & 1.000 & deposit $\rightarrow$ put & 0.872 & 0.989\\
grapes $\rightarrow$ wine & 0.931 & 1.000 & central $\rightarrow$ middle & 0.911 & 0.994\\
citrus $\rightarrow$ orange & 0.943 & 0.995 & uppermost $\rightarrow$ top & 0.893 & 0.973\\
\bottomrule
\end{tabular}
\caption{\textbf{Cosine similarity to the destination embedding.} Values are recomputed from the original and edited row vectors; each row is one actual tokenizer piece. The exact token IDs and all 65 mappings are listed below and in the accompanying JSON.}
\label{tab:alignment}
\end{table*}
\begin{table*}[t]
\centering\scriptsize
\begin{minipage}[t]{.485\textwidth}\centering
\setlength{\tabcolsep}{2pt}\begin{tabular}{rp{.72in}p{1.23in}r}\toprule ID & Source & Destination(s) & $\alpha$\\\midrule
1201 & round & bowl & 0.50\\
2851 & position & put & 0.75\\
3425 & \_cook & \_stove & 0.75\\
3668 & \_position & \_put & 0.75\\
4871 & \_higher & \_top & 0.75\\
5094 & \_round & \_bowl & 0.50\\
5880 & framework & rack & 0.75\\
6985 & \_flat & \_plate & 0.50\\
7233 & \_central & \_middle & 0.75\\
7825 & \_pull & \_open & 0.75\\
8151 & \_spread & \_cream, \_cheese, \_butter & 0.75\\
8776 & \_upper & \_top & 0.75\\
8831 & insert & put & 0.75\\
9393 & \_highest & \_top & 0.75\\
9499 & \_slide & \_push & 0.75\\
10548 & \_secure & \_close & 0.75\\
14455 & \_insert & \_put & 0.75\\
15087 & \_framework & \_rack & 0.75\\
15476 & \_grab & \_pick & 0.75\\
15694 & \_deposit & \_put & 0.75\\
15983 & \_cooking & \_stove & 0.75\\
16036 & \_drinking & \_mug & 0.75\\
19098 & flat & plate & 0.50\\
23801 & pull & open & 0.75\\
23832 & upper & top & 0.75\\
23865 & slide & push & 0.75\\
25373 & \_mixing & \_bowl & 0.75\\
26836 & \_rounded & \_bowl & 0.75\\
30506 & \_grasp & \_pick & 0.75\\
31062 & \_dairy & \_cheese, \_milk, \_butter & 0.50\\
31725 & \_basin & \_bowl & 1.00\\
32006 & central & middle & 0.75\\
32518 & \_elevated & \_top & 0.75\\
\bottomrule\end{tabular}\end{minipage}\hfill
\begin{minipage}[t]{.485\textwidth}\centering
\setlength{\tabcolsep}{2pt}\begin{tabular}{rp{.72in}p{1.23in}r}\toprule ID & Source & Destination(s) & $\alpha$\\\midrule
33182 & secure & close & 0.75\\
33773 & \_activate & \_turn & 0.75\\
37806 & activate & turn & 0.75\\
42226 & \_grape & \_wine & 1.00\\
46405 & spread & cream, cheese, butter & 0.75\\
48841 & rounded & bowl & 0.75\\
50161 & \_grapes & \_wine & 1.00\\
57321 & \_citrus & \_orange & 0.75\\
57655 & \_broth & \_soup & 0.75\\
58983 & deposit & put & 0.75\\
59031 & grab & pick & 0.75\\
65931 & \_cooker & \_stove & 1.00\\
69707 & highest & top & 0.75\\
70624 & \_cultured & \_cheese & 0.75\\
73111 & higher & top & 0.75\\
80792 & \_forefront & \_front & 0.75\\
81247 & \_atop & \_top & 0.75\\
82203 & \_cupboard & \_cabinet & 1.00\\
96159 & \_churn & \_butter & 0.75\\
110185 & \_relocate & \_put & 0.75\\
114035 & atop & top & 0.75\\
118891 & \_fermented & \_wine & 1.00\\
121144 & \_ignite & \_turn & 0.75\\
125367 & basin & bowl & 1.00\\
135139 & \_shove & \_push & 0.75\\
136152 & mixing & bowl & 0.75\\
139848 & drinking & mug & 0.75\\
157695 & grape & wine & 1.00\\
160148 & \_uppermost & \_top & 0.75\\
167690 & ignite & turn & 0.75\\
177466 & dairy & cheese, milk, butter & 0.50\\
204615 & citrus & orange & 0.75\\
\bottomrule\end{tabular}\end{minipage}
\caption{All 65 edited rows in C. The complete machine-readable recipe and original/edited vectors are included in the reproducibility package.}\label{tab:allrows}\end{table*}

\clearpage
\onecolumn
\section{Qualitative Examples and Task Catalogs}
\label{app:qualitative}
The following five pages show archived simulator observations: four matched held-out $\pi_{0.5}$ repairs, four original-weight FastWAM input-format pairs, and 22 practice task images. All held-out video action hashes match their evaluator records. Images show the agent camera; $t$ is a recorded policy step, with video frame $j$ corresponding to $2j$. Last frames can precede terminal checks. The figure captions specify selection and whether a sample is practice or held-out. All four C regressions are retained in Table~\ref{tab:regression-prompts}. Source manifests give exact prompts, episode IDs, seeds, outcomes, and hashes.

\begin{figure}[!ht]
\centering\includegraphics[width=\textwidth]{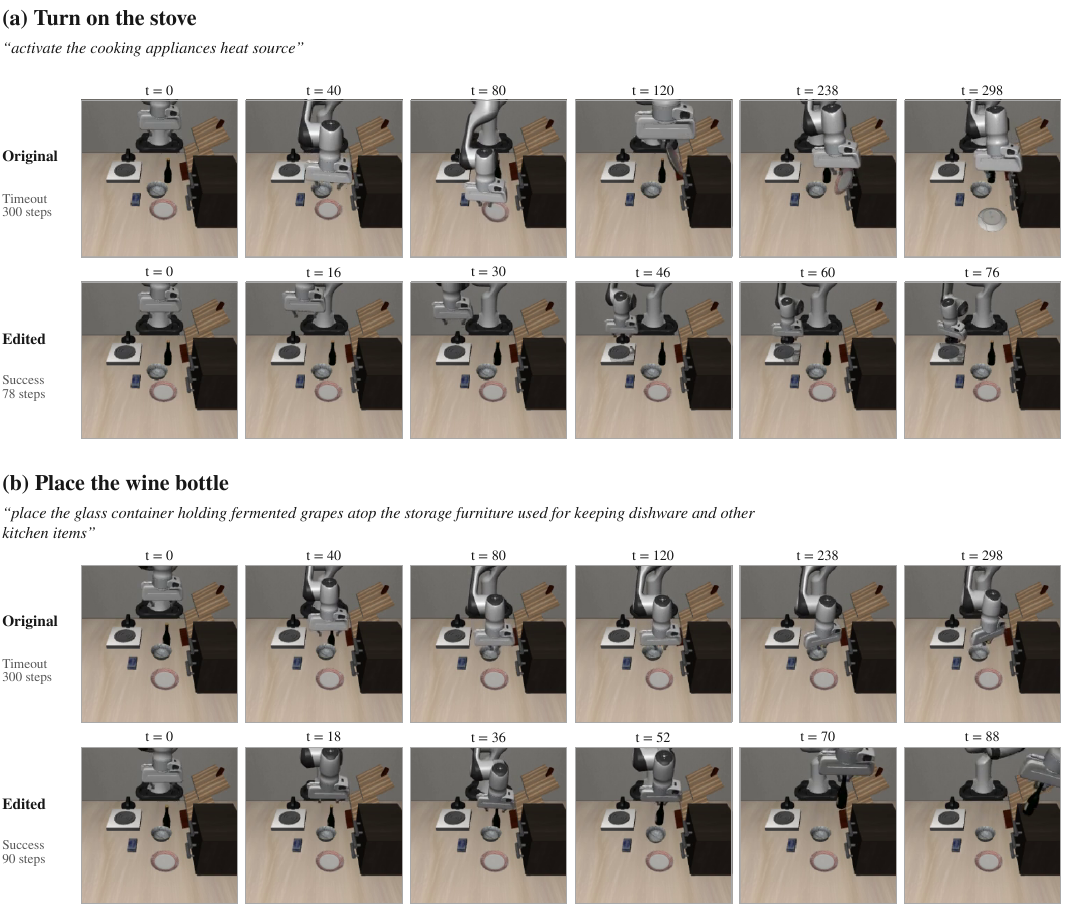}
\caption{\textbf{Held-out language repairs: stove and wine placement.} Six recorded observations per trajectory for LIBERO-Goal variants 1421 and 1198, seed 7, initial state 0. Original and edited weights receive the same instruction. Original episodes time out at 300 steps; edited episodes succeed at 78 and 90. All $t$ labels identify recorded policy steps, not normalized video time. These are two of C's 48 repairs.}
\label{fig:supp-pi-a}
\end{figure}

\clearpage
\begin{figure}[!ht]
\centering\includegraphics[width=\textwidth]{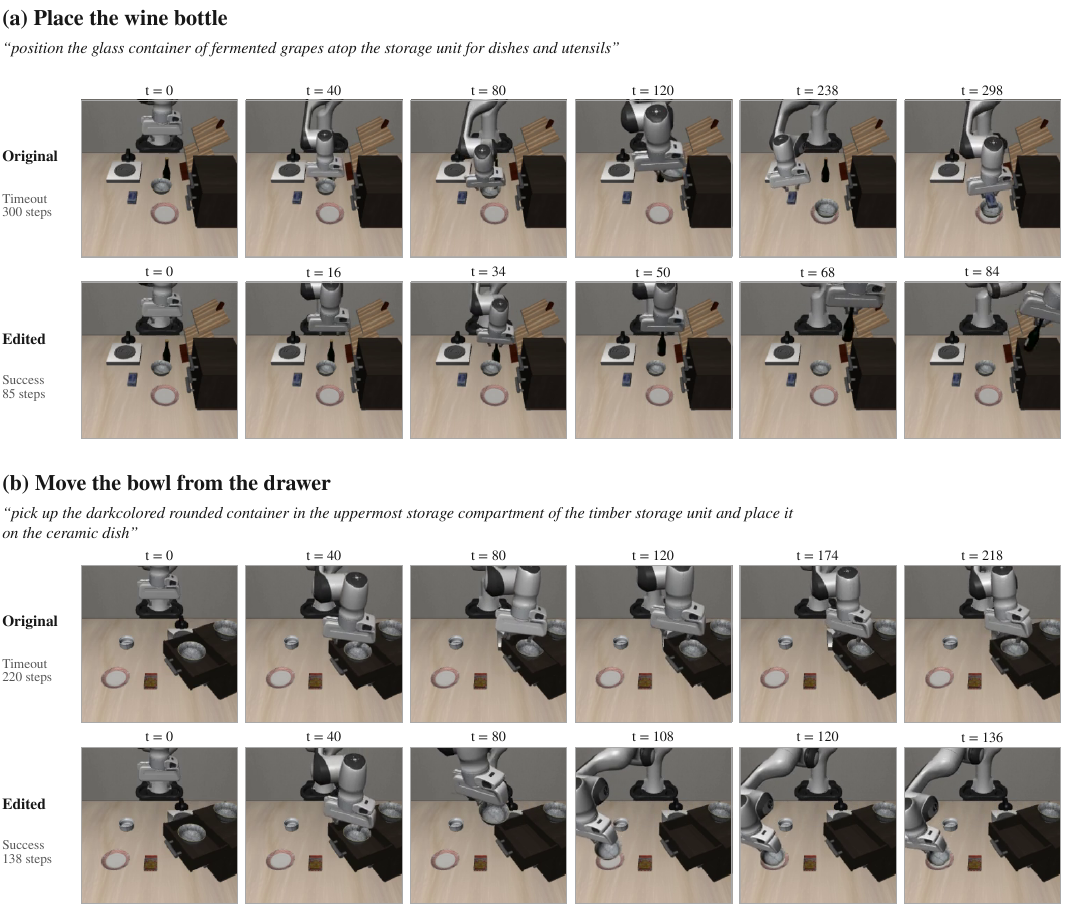}
\caption{\textbf{Held-out language repairs: descriptive wine and drawer instructions.} LIBERO-Goal 1174 and LIBERO-Spatial 1148, seed 7, initial state 0. Original episodes time out at 300 and 220 steps; edited episodes succeed at 85 and 138. Different instructions for the same wine task and a different manipulation task provide additional views of the lexical repair. These remain selected successful examples; the four held-out regressions are listed in Table~\ref{tab:regression-prompts}.}
\label{fig:supp-pi-b}
\end{figure}

\clearpage
\begin{figure}[!ht]
\centering\includegraphics[width=\textwidth]{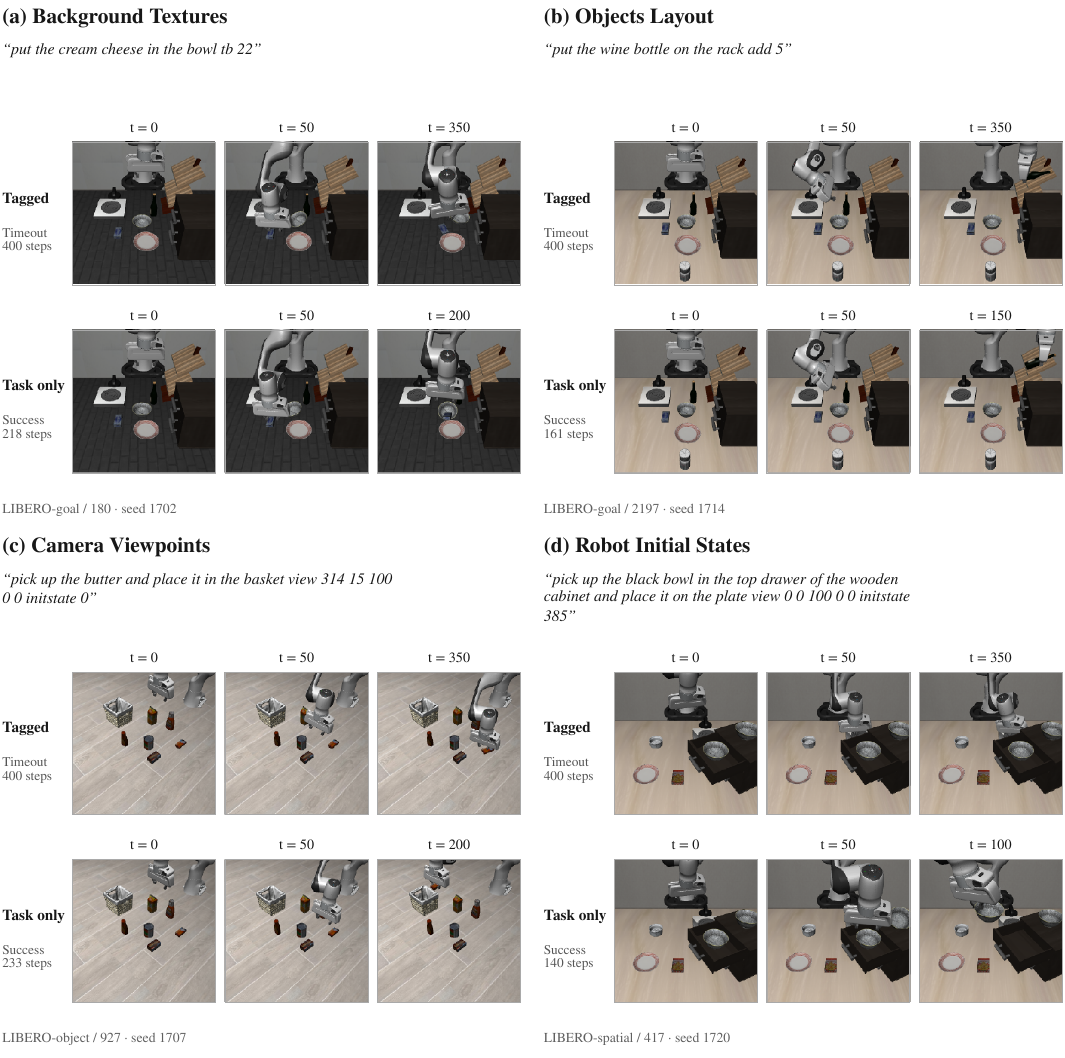}
\caption{\textbf{FastWAM under tagged and task-only instructions, original weights in both rows.} Four practice variants from the background, camera, object-layout, and initial-state categories. Each panel prints the exact tagged instruction and the episode ID/seed. ``Task only'' uses the corresponding recorded task-only string, with metadata removed. First, intermediate, and last saved observations are shown; the last is not necessarily the terminal state because these diagnostics save every 50 steps. The source manifest contains both full strings and terminal records. The four outcomes change from failure to success, without any weight edit.}
\label{fig:supp-fw-format}
\end{figure}

\clearpage
\begin{figure}[!ht]
\centering\includegraphics[width=\textwidth]{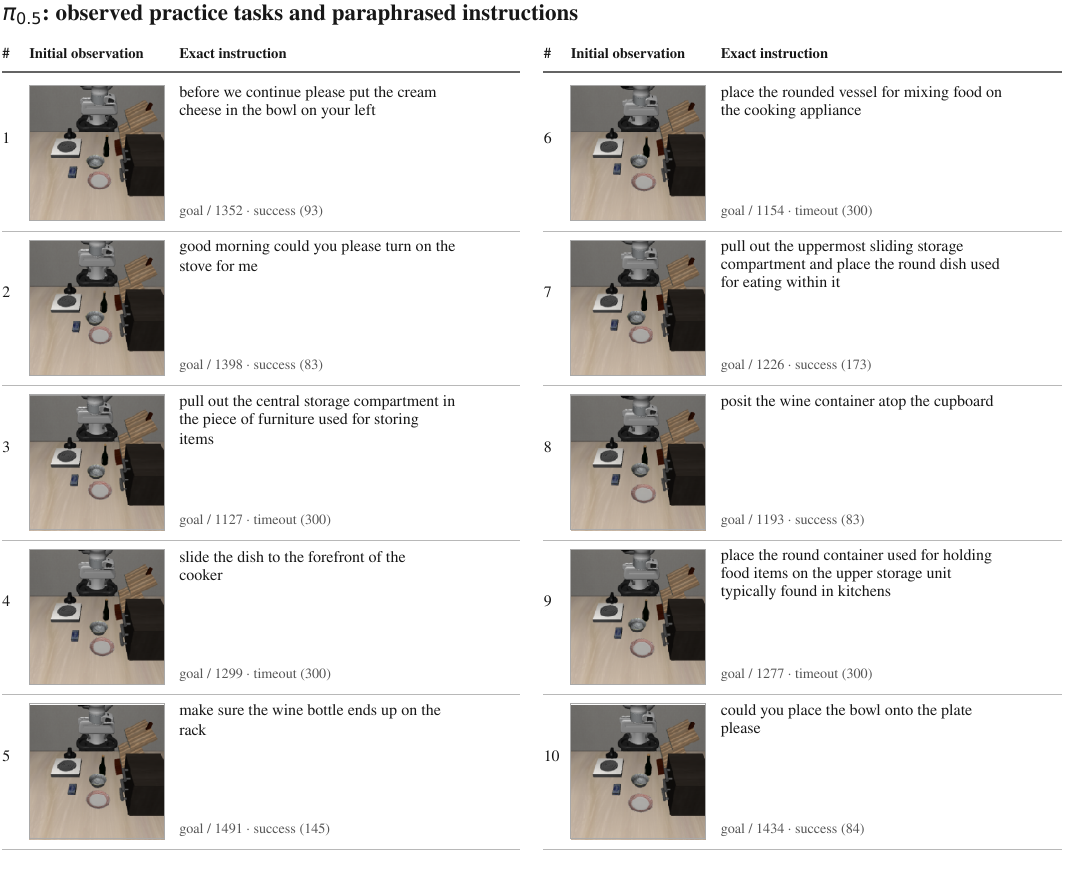}
\caption{\textbf{Ten observed LIBERO-Goal practice tasks for $\pi_{0.5}$.} Each row pairs the initial agent-camera image with the exact paraphrased instruction. Metadata below each prompt gives suite/variant ID and original-policy outcome with step count. These rows describe the practice scan; they are not ten additional tests of the final edit. All ten task prefixes are included once in their first-occurrence order.}
\label{fig:catalog-pi}
\end{figure}

\clearpage
\begin{figure}[!ht]
\centering\includegraphics[width=\textwidth]{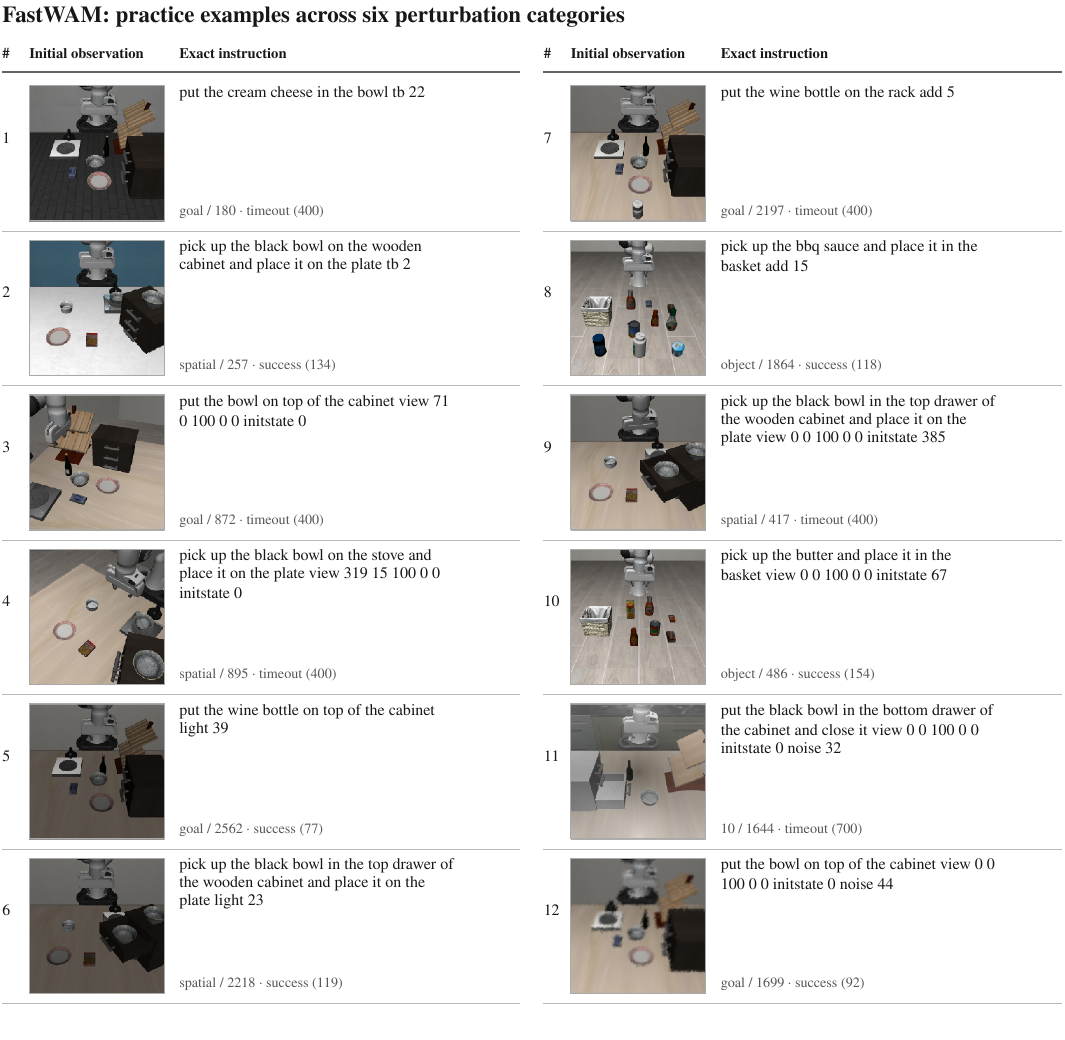}
\caption{\textbf{FastWAM practice examples across all six non-language perturbation categories.} The 12 initial observations retain the camera, background, object-layout, lighting, robot-state, and sensor-noise variation in the archived diagnostic. Exact tagged prompts and original-policy outcomes are printed next to the images. Rows are grouped by category: 1--2 background, 3--4 camera, 5--6 lighting, 7--8 layout, 9--10 robot state, and 11--12 noise. These are observed simulator images, not generated video predictions.}
\label{fig:catalog-fw}
\end{figure}

\end{document}